\documentclass{article}

\usepackage{PRIMEarxiv}

\usepackage[utf8]{inputenc} 
\usepackage[T1]{fontenc}    
\usepackage{hyperref}       
\usepackage{url}            
\usepackage{booktabs}       
\usepackage{amsfonts}       
\usepackage{nicefrac}       
\usepackage{microtype}      
\usepackage{lipsum}
\usepackage{fancyhdr}       
\usepackage{graphicx}       
\graphicspath{{media/}}     
\usepackage{fontawesome}
\usepackage{subfigure}
\usepackage{multirow}
\usepackage{multicol}
\title{Aquila2 Technical Report
}

\author{
  {\bf Bo-Wen Zhang}, {\bf Liangdong Wang}, {\bf Jijie Li}, {\bf Shuhao Gu}, {\bf Xinya Wu}, \\
{\bf Zhengduo Zhang}, {\bf Boyan Gao}, {\bf Yulong Ao}, {\bf Guang Liu}\thanks{*Project lead, the corresponding author, \url{liuguang@baai.ac.cn}} \\
        Language Foundation Model \& Software Team \\ BAAI 
}

\begin{document}
\maketitle

\begin{abstract}
This paper introduces the Aquila2 series, which comprises a wide range of bilingual models with parameter sizes of 7, 34, and 70 billion.
These models are trained based on an innovative framework named HeuriMentor (HM), which offers real-time insights into model convergence and enhances the training process and data management. The HM System, comprising the Adaptive Training Engine (ATE), Training State Monitor (TSM), and Data Management Unit (DMU), allows for precise monitoring of the model's training progress and enables efficient optimization of data distribution, thereby enhancing training effectiveness. Extensive evaluations show that the Aquila2 model series performs comparably well on both English and Chinese benchmarks. Specifically, Aquila2-34B demonstrates only a slight decrease in performance when quantized to Int4. Furthermore, we have made our training code\footnote{https://github.com/FlagOpen/FlagScale} and model weights\footnote{https://github.com/FlagAI-Open/Aquila2} publicly available to support ongoing research and the development of applications.
\end{abstract}


\section{Introduction}

Large Language Models (LLMs) exhibit remarkable competence in a wide range of downstream tasks and are catalyzing a fundamental shift in research paradigms~\cite{DBLP:conf/nips/BrownMRSKDNSSAA20,chatgpt,openai2023gpt4}. Data play a crucial role in the model training process. Lately, there has been significant focus on investigating the effects of different training data combinations, such as OPT~\cite{zhang2022opt}, Bloom~\cite{DBLP:journals/corr/abs-2211-05100}, Palm~\cite{chowdhery2022palm}, and LLaMA~\cite{DBLP:journals/corr/abs-2302-13971}. These models are typically trained on static datasets over long periods. For instance, LLaMA 65B underwent a 21-day training session in 2048 A100 GPU with 80GB of RAM~\cite{touvron2023llama}. Nonetheless, conventional training methods frequently struggle to adapt to variations in data composition or the integration of new data. Due to the resource-intensive nature of each training iteration, enhancing training approaches is crucial for the effective training of LLMs.

This paper introduces the Aquila2 series, which comprises bilingual models ranging in parameter sizes from 7 to 70 billion. The HeuriMentor (HM) Framework is developed to improve the training efficiency of the Aquila series models. The HM System consists of the Adaptive Training Engine (ATE), the Training State Monitor (TSM), and the Data Management Unit (DMU). By integrating these components, the system allows for better monitoring of the model's training progress and enables efficient adjustments to the data distribution for optimizing training effectiveness. The Adaptive Training Engine (ATE) is designed to train models by continuously updating a mixture of data that includes the most recent data sources, thus enhancing model performance on subsequent tasks. ATE allows for the flexible modification of cluster sizes during training, such as moving from a 12xA100 40G cluster to a 16xA800 80G cluster. Moreover, ATE facilitates training on diverse devices. The Data Management Unit (DMU) is in charge of gathering and organizing data from the Internet and collaborators to use for model training. Initially, the DMU collects information from web pages and PDFs, then proceeds with thorough deduplication and quality filtering. Moreover, the DMU oversees the data combination recipe for each training cycle. We have explored various combinations of these diverse data sources, detailing the outcomes of these experiments to offer empirical insights and key learnings. The Training State Monitor (TSM) is designed to evaluate the development of models trained by the Adaptive Training Engine in real time. By tracking metrics such as loss, downstream performance, and model weight changes throughout the training process, we can seamlessly incorporate new data or adjust data combinations as needed. This approach sets up a continuous learning feedback loop that enables the system to improve its performance by leveraging insights from past results.

We conducted a thorough evaluation of the Aquila2 Series using various benchmarks. The extensive experimental findings indicate that Aquila2-34B consistently outperforms the baseline models in terms of average scores. This suggests that the Aquila2-34B model shows improved performance with fewer training tokens compared to LLaMA-2-70B and other bilingual models tested across 21 diverse datasets. Importantly, we noticed only minimal performance degradation in Aquila2-34B when subjected to 4-bit quantization. Besides, we collected instructional data to train the chat version of the Aquila2 models. The evaluation of AquilaChat2-34B using a bilingual benchmark, which included subjective and objective assessments, consistently illustrates the superior performance of AquilaChat2-34B over LLaMA-2-70B and its corresponding chat models.


\section{Aquila2 series}

\textbf{Tokenizer.} To identify the optimal vocabulary size, preliminary experiments were carried out with varying vocabulary sizes and language proportions. Based on the results, we decided to set Aquila2's vocabulary size at 100,000. Simultaneously, we adopted Byte Pair Encoding (BPE)~\cite{DBLP:conf/acl/SennrichHB16a} to extract the vocabulary with huggingface transformers~\cite{DBLP:journals/corr/abs-1910-03771} tool. The training data corpus consists of an equal proportion of English and Chinese, sourced from WudaoCorpus~\cite{DBLP:journals/aiopen/YuanZDDLCZYT21} and Pile~\cite{gao2020pile} for Chinese and English data, respectively. 

\textbf{Group Query Attention.} We adopt the Grouped Query Attention (GQA) ~\cite{ainslie2023gqa} mechanism, which demonstrates an enhanced level of efficiency in comparison to traditional multi-head attention during the inference process. Previous work~\cite{touvron2023llama} illustrates that pre-trained GQA models produce quality levels that closely match those of conventional multi-head attention models while maintaining processing speeds that are comparable to those of the MHA models. 

\textbf{Position Embedding.} In the architecture of Aquila2, we deploy Rotary Position Embedding (RoPE)~\cite{DBLP:journals/corr/abs-2104-09864}, a widely recognized and adopted methodology within the paradigm of large language model architectures. The incorporation of RoPE, which ingeniously amalgamates the benefits of both traditional relative and absolute position encoding, aims to enhance the potency of the model by efficiently capturing the underlying spatiotemporal patterns in sequence data.

\begin{table*}[t]
\centering
\setlength{\tabcolsep}{1mm}
\begin{tabular}{lcccccccc}\hline
Model & Layers & Hidden  & FFN  &  Head & GQA  & MaxLen & LR & Batch Size  \\\hline
Aquila2-7B   & 32  &    4096   & 11008   &       32                  &      -                     &        2048        & 2e-4 &  1728     \\
Aquila2-34B  & 60  & 6144     & 24576 &       48                  &      8                       &           4096  & 1.5e-4 & 1024 \\
Aquila2-70B-expr & 80 & 8192 & 28672& 64 & 8 & 4096 & 1.5e-4& 1032\\\hline
\end{tabular}
\caption{The training configurations for the Aquila2 series models. }
\label{tab:structure}
\end{table*}

Other details of the model structure are listed in Table~\ref{tab:structure}.

\section{HeuriMentor Framework}
\begin{figure*}[bhtp]
    \centering
    \includegraphics[width=0.9\columnwidth]{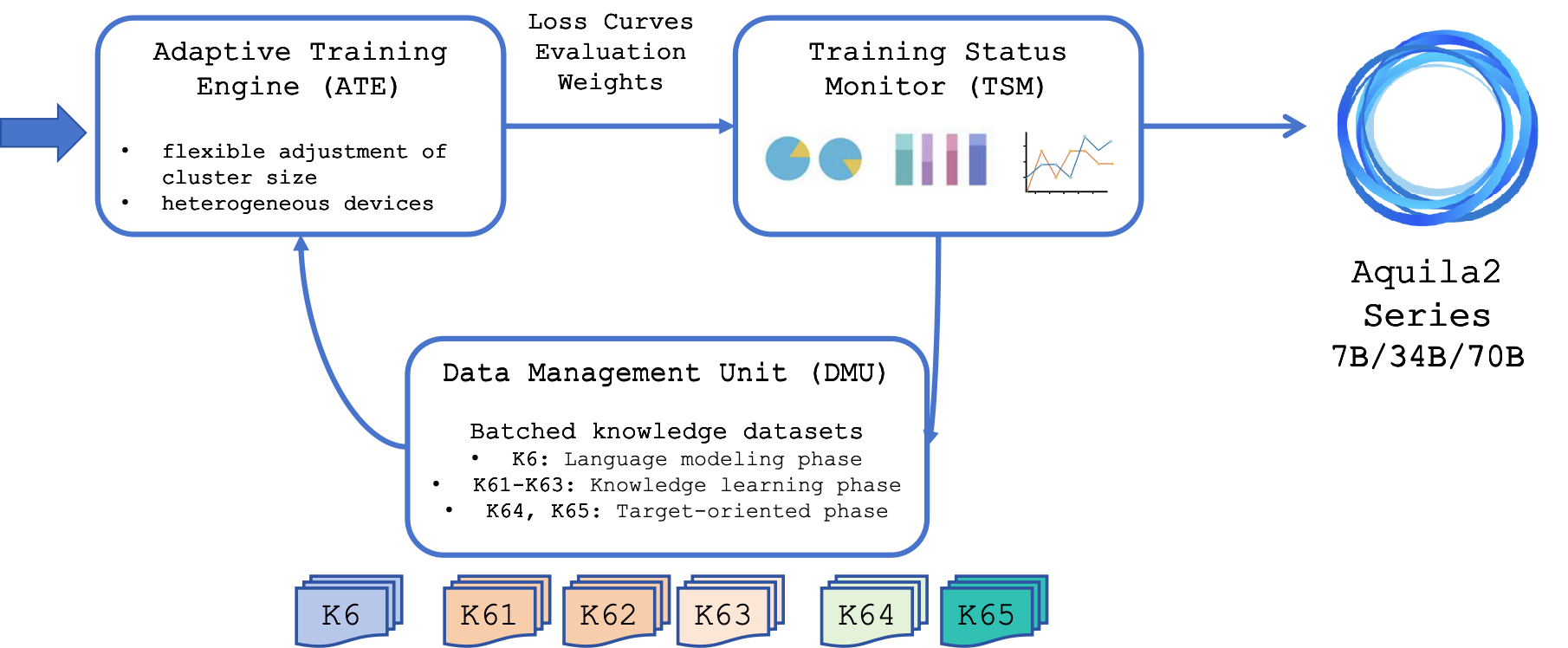}
    \caption{The HeuriMentor Framework structure. }
    \label{fig:HM_framework}
\end{figure*}

The HeuriMentor framework focuses on training models effectively with dynamically changing data mixtures. In Figure~\ref{fig:HM_framework}, two main scenarios are illustrated: first, when a new data source is acquired, and second, when the goal is to improve the model's performance for downstream tasks.
\subsection{Adaptive Training Engine} 
The Adaptive Training Engine (ATE) is designed to train models using an updated mix of data from the latest sources. This training approach aims to enhance the performance of models on subsequent tasks. To ease the need for computational resources, ATE accommodates various model structures and parameter scales, with the training strategy optimized across multiple chips. ATE also allows for flexible adjustment of cluster sizes during training. Importantly, ATE supports training on heterogeneous devices (Aquila2-70B-expr is trained on heterogeneous devices).

In the training of Aquila2-7B, we employ data parallelism and a distributed optimizer to enhance efficiency \cite{ rajbhandari_zero_2019,li_pytorch_2020}. For Aquila2-34B, to address its significant GPU memory demands without impacting GPU utilization, we incorporate data and tensor parallelism, coupled with a distributed optimizer and integrated with 1F1B pipeline and sequence parallelism \cite{shoeybi_megatron-lm_2020,narayanan_efficient_2021,korthikanti_reducing_2022}. Remarkably, Aquila2-70B-Expr, trained under a heterogeneous strategy, is an experimental version that operates on a mixed cluster of A100 40G and A800 80G GPUs, employing specifically optimized strategies for the heterogeneous devices. Moreover, Aquila2 utilizes FlashAttion-2~\cite{DBLP:journals/corr/abs-2307-08691} to further increase computational efficiency. This approach results in roughly 666 tokens/sec/GPU during Aquila2-34B's training.

\textbf{Mixed precision.} We utilize bfloat16 in model training for numerical stability, supplemented by float32 for precision-critical operations, encompassing word and Rotary position embeddings, attention softmax, and gradient all-reduce.

\textbf{Hyperparameters.} In the preliminary training phase encompassing the initial 8B tokens, the batch size is progressively elevated from 32 to 1024. The AdamW optimizer is employed with the hyperparameters set as follows: $\beta_{1} = 0.9$, $\beta_{2} = 0.95$, $eps = 10^{-8}$, complemented by a weight decay of $0.1$. Our approach incorporates a cosine learning rate schedule and imposes gradient clipping at $1.0$. The learning rate undergoes an initial escalation to $1.5\times10^{-4}$ within the first 2B tokens, then decays to $1.5\times10^{-5}$. Dropout techniques were not applied during the training process.

\textbf{Training hardware.} The Aquila2-7B model is pre-trained on 11x8 NVIDIA A800-80GB GPUs, while the Aquila2-34B model is pre-trained on 64x8 NVIDIA A100-40GB GPUs. Both clusters feature an interconnect equipped with 2x200 Gbps InfiniBand. The A800 cluster has an intra-node GPU connection of 400 GB/s, whereas the A100 cluster offers 600 GB/s.

\subsection{Training State Monitor (TSM)}
Our Training State Monitor is designed to analyze the status of models trained by the Adaptive Training Engine dynamically. During training, we track metrics such as loss, downstream performance, and model weights to monitor the model's status. This allows us to purposefully incorporate new data or adjust the data mixture as required. This approach introduces a dynamic learning feedback loop, enabling the system to improve its performance by incorporating insights from prior results. If the subsequent Language Model (LLM) fails to surpass its predecessor, adjustments can be made to the training procedures or the data used for training to achieve optimal efficiency. Furthermore, this framework provides a versatile tool that effectively enhances the performance of LLMs across a wide range of tasks, especially those requiring intensive knowledge.

In the initial phase, our Aquila2 models are fine-tuned using a well-researched data distribution method~\cite{hoffmann2022training, touvron2023llama}. This stage involves exposing the models to approximately 200-300 billion tokens, which facilitates the model to generate predictions based on a minimal collection of exemplars.

After training the models, the next step is to monitor their performance. This involves recording and assessing any changes in efficiency, detecting patterns, and highlighting deviations. By closely monitoring the models, potential issues can be detected and corrected early.

In the final phase, the data distribution is adjusted based on empirical evidence. Some possible adjustments include increasing the sampling of under-represented classes, readjusting the data distribution, or introducing data changes to elicit diverse responses from the model. The goal of this iterative redistribution process is to maximize the overall performance of the model. Later sections will provide more details on refining this method and explaining its potential applications in a wider context.
\begin{figure}[hbpt]
\centering
\includegraphics[width=0.7\columnwidth]{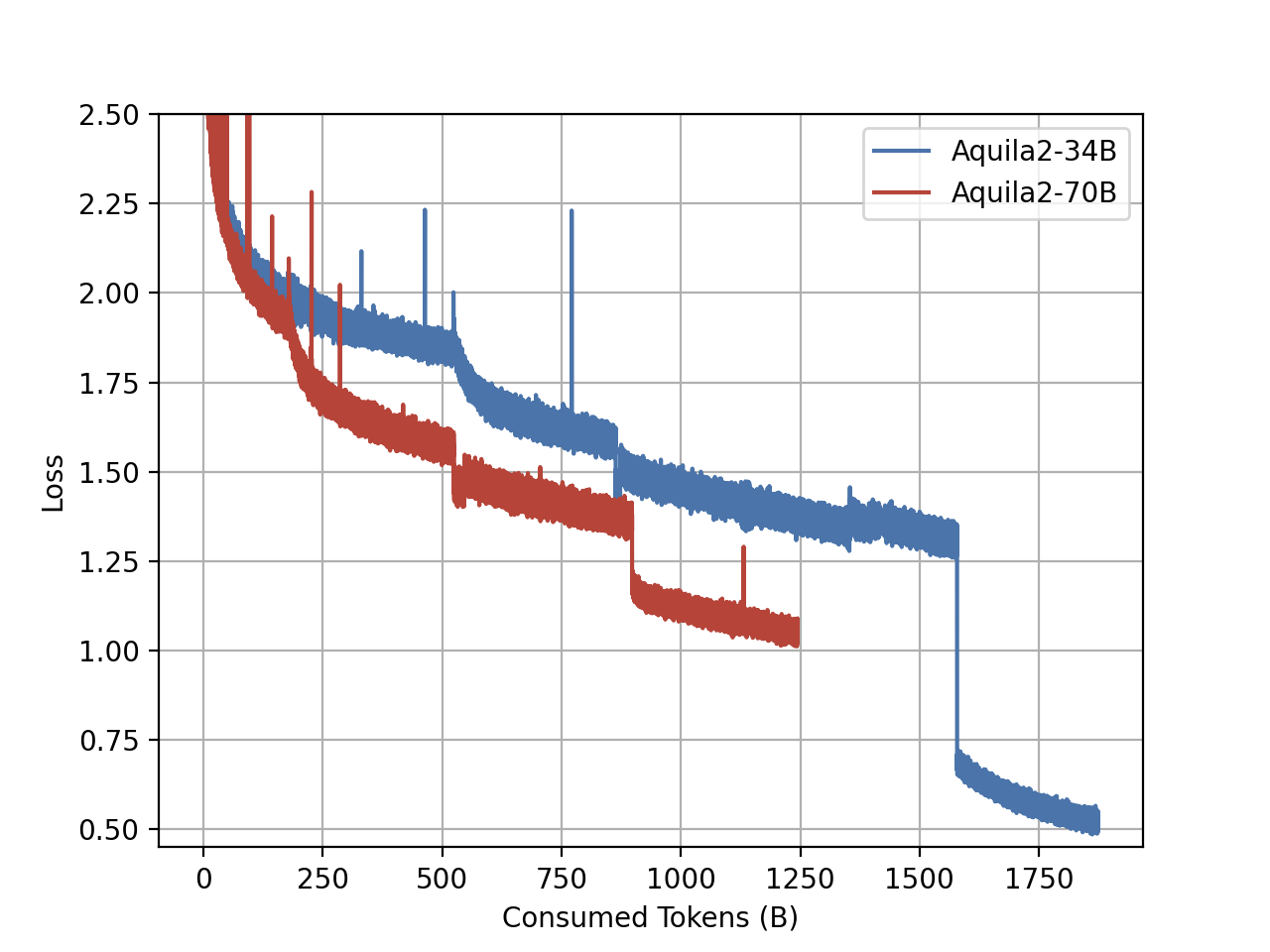}
\caption{Training loss for Aquila2-34B and Aquila2-70B-expr Models\label{loss}.}
\end{figure}
\subsubsection{Training loss} 
The training loss curves of the Aquila2-34B and Aquila2-70B-expr models, as shown in Figure~\ref{loss}, reveal key insights into the impact of strategic data adjustments on model performance. Initially, both models exhibit a steady decline in loss, which gradually slows as the training progresses. This flattening of the loss curve is indicative of the models reaching a phase where further improvements become more challenging, likely due to the models approaching a local minimum or the limits of what can be learned from the current dataset.

Recognizing this trend, adjustments were made to the training process, such as introducing new data, modifying the learning rate, or other forms of data augmentation. These adjustments are reflected in the sharp drops in the loss curves following the plateaus. These interventions effectively reinvigorate the training process, leading to a marked improvement in training efficiency. This is evident from the more pronounced and rapid decreases in loss after each adjustment.

The Aquila2-70B-expr model, in particular, shows a consistently lower loss compared to the Aquila2-34B model, demonstrating its superior capacity to learn and generalize from the data. This can be attributed to its larger parameter space, which allows it to model more complex relationships within the data. The observed patterns highlight the importance of strategically timing these interventions to overcome stagnation in training and to enhance the overall performance of the models.

In the early training stages, we observed unexpected spikes in the training loss. This phenomenon occurred more frequently in Aquila2-70B than in Aquila2-34B and was attributed to a low-quality, noisy dataset in certain language topics. To stabilize training without rolling back to previous checkpoints, we meticulously cleaned the dataset before feeding it into the model. We monitored the norm of the model weights, assessing their performance on downstream tasks to ensure smooth training. As a result, the frequency of these spikes significantly decreased.


A sharp decrease in training loss is observed in Aquila2-34B after approximately 1500 tokens. However, this decrease does not directly indicate an improvement in model performance, as there are no immediate differences in the downstream tasks. Instead, we attribute this to the characteristics of the training data, specifically the knowledge-specific dataset used at that point in the training process.

\begin{figure}[t]
\centering
\begin{subfigure}[]{\includegraphics[width=0.45\linewidth]{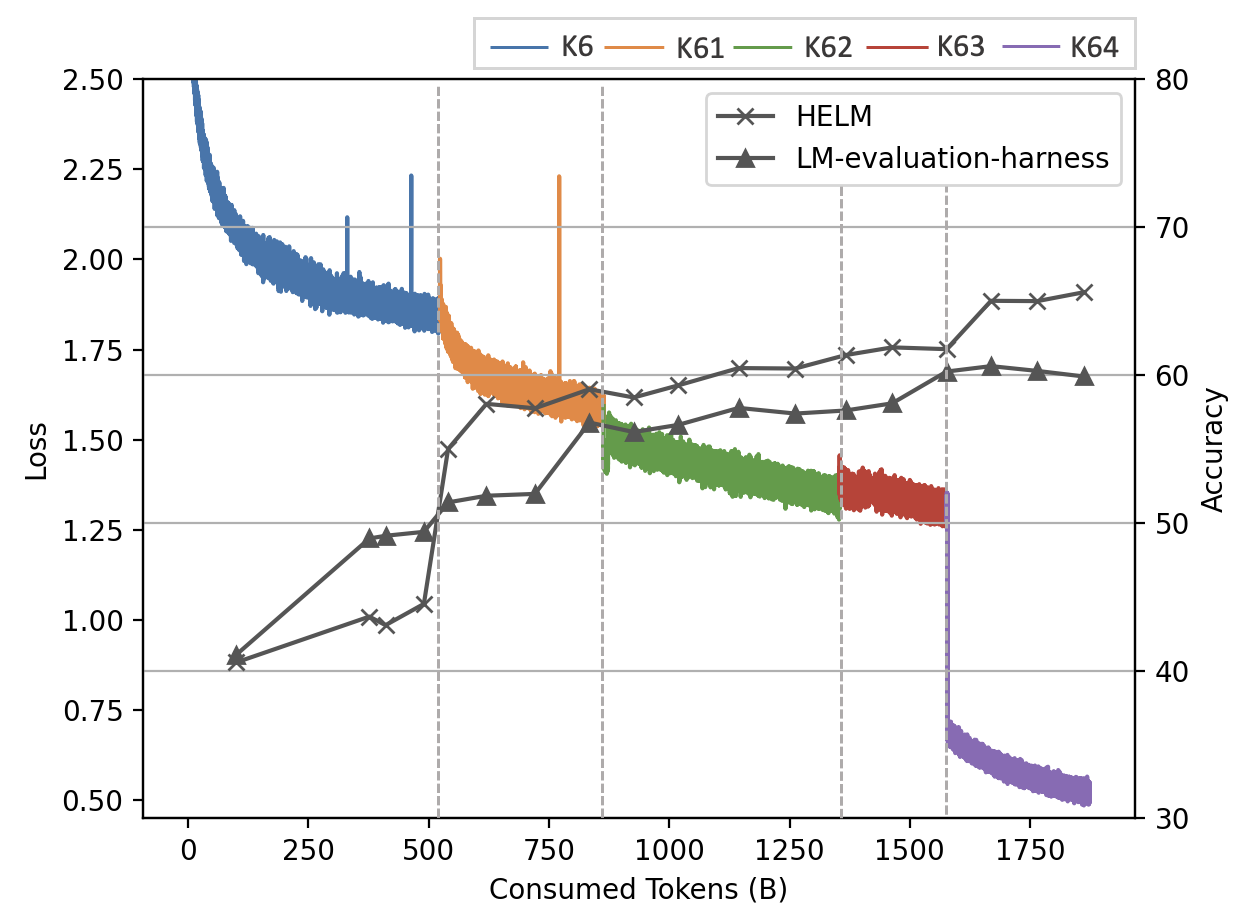}}
\end{subfigure}
\hfill
\begin{subfigure}[]{\includegraphics[width=0.45\linewidth]{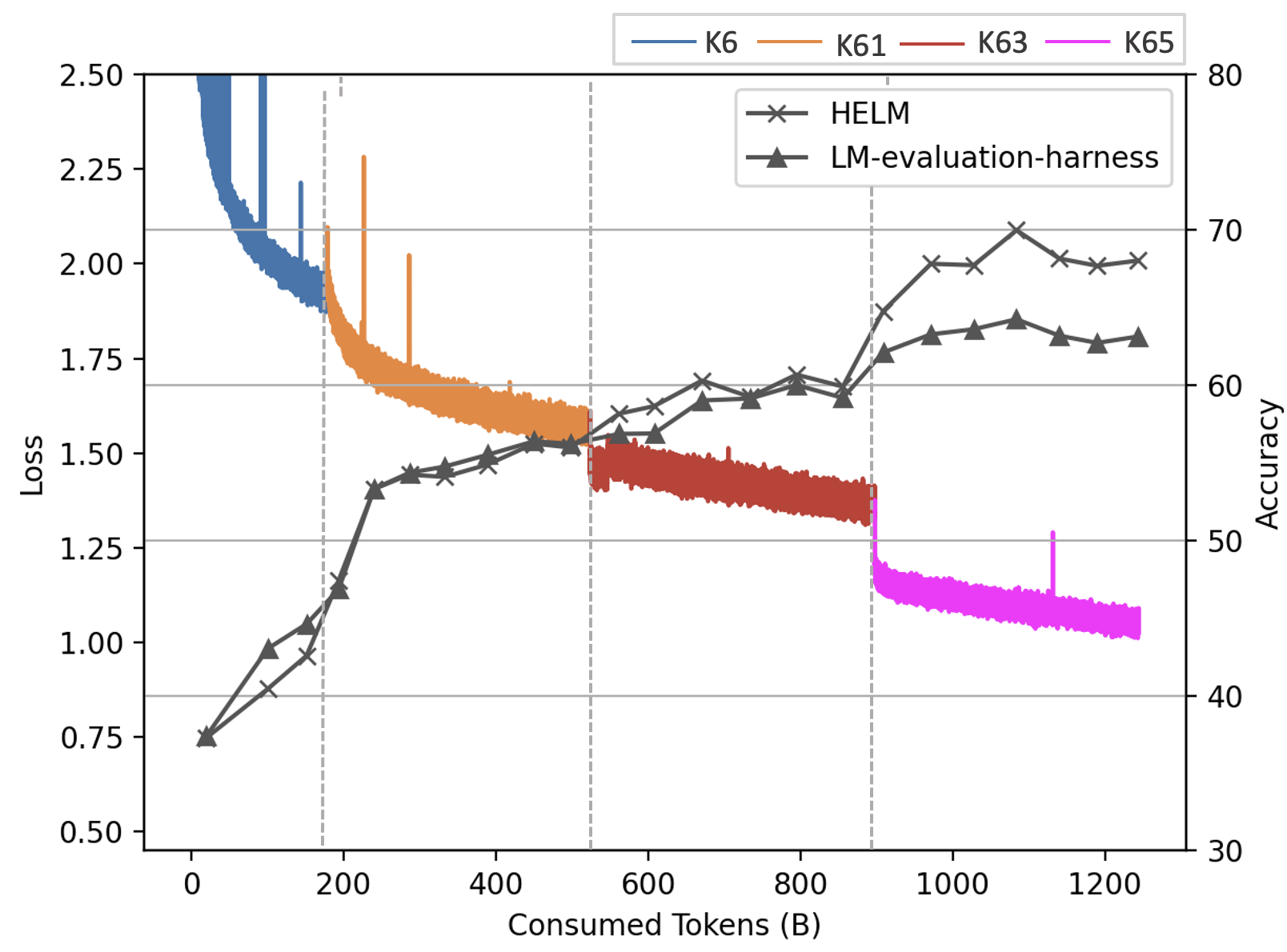}}
\end{subfigure}
\caption{Performance of Aquila-34B (a) and Aquila-70B-expr (b) on downstream tasks during training. We use different colors to distinguish between different data stages of the training loss. We apply the score of HELM and LM evaluation as the main metrics. Details of evaluation are covered in section \ref{evaluation metrics}.}
\label{downstream-performance}
\end{figure}

\subsubsection{Downstream performance} 

In Figure~\ref{downstream-performance}, we observe the training performance of two large language models, Aquila-34B (a) and Aquila-70B-expr (b), as represented by their loss and accuracy metrics over consumed tokens. The loss curves are initially marked by distinct stages of training, represented by different colors, which indicate variations in the data employed at each stage.

For Aquila-34B (a), the loss decreases rapidly during the initial stages, indicating effective learning. However, as training progresses and the model stabilizes, the rate of loss reduction becomes more gradual. At this point, adjustments in the data (as indicated by the color changes) are introduced to address the plateau in performance. This intervention leads to a marked improvement in training efficiency, as evidenced by a sharper decline in the loss and corresponding improvements in accuracy during subsequent stages.

Similarly, in Aquila-70B-expr (b), we observe a comparable trend where the loss reduction slows down after initial rapid learning. The introduction of data adjustments (denoted by color shifts) again leads to significant gains in model performance. The accuracy, which corresponds inversely to the loss, also demonstrates notable improvements post-adjustment, reflecting the model's enhanced ability to generalize from the modified training data.



\begin{figure}[t]
\centering
\begin{subfigure}[]{\includegraphics[width=0.45\linewidth]{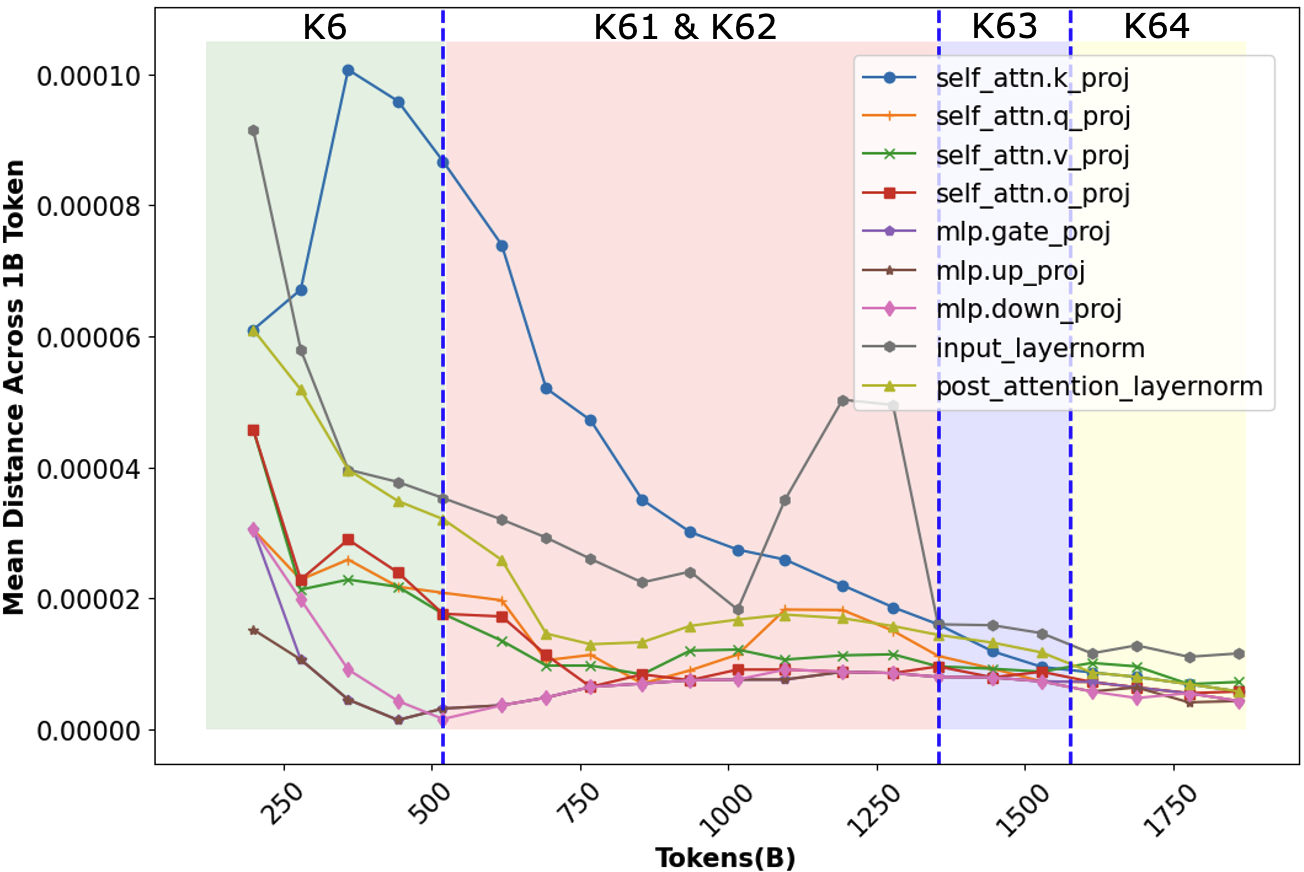}}
\label{fig:weights_34B}
\end{subfigure}
\hfill
\begin{subfigure}[]{\includegraphics[width=0.45\linewidth]{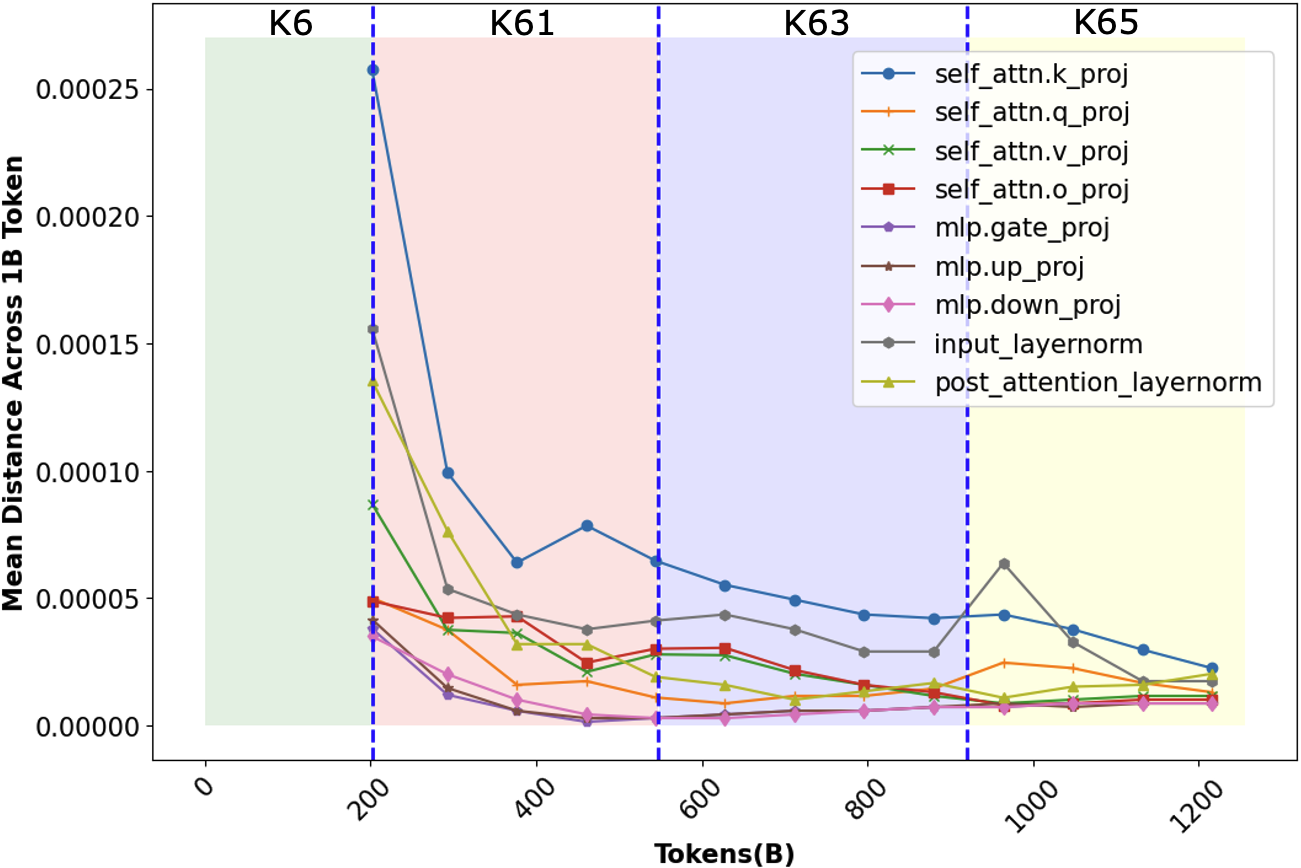}}
\label{fig:weights_70B}
\end{subfigure}
\caption{Convergence evidence from the perspective of the weight. Different colors of rectangles represent different data stages, corresponding to the stages(K6, K61\&K62, K63, and K64) in Aquila2-34B, and the stages(K6, K61, K63, and K65) in Aquila-70B.\label{parameter}}
\end{figure}

\subsubsection{Weight Trajectory}
Figure.~\ref{parameter} presents the convergence behavior of different weight matrices across training stages for both Aquila-34B (a) and Aquila-70B (b) models. The x-axis represents the number of tokens processed, while the y-axis indicates the mean distance between weights across a billion tokens. The colored rectangles demarcate distinct training stages. Detailed parameter distributions are shown in Appendix.\ref{parameter_observation}

This section explores the effects of the different batches of pretraining datasets on the convergence of the Aquila2 model. We evaluated this by monitoring the trajectory of model parameter distribution during training, which is visually depicted in Figure.~\ref{parameter} using the standard deviation of each parameter across transformer layers. For this analysis, $N$ is the number of transformer layers and $M$ is the number of checkpoints, with the parameter for the layer $k$ at the checkpoint $i$ referred to as $layer_i^k$. We focus on nine essential parameters of $layer_i^{k,p}$, where $p$ refers to specific features of the model. The standard deviation of $layer_i^{k,p}$ is signified as $std_i^{k,p}$. Given that the average value of each parameter is fixed at 0, our primary metric is the standard deviation as it reflects weight alterations. $std_i^{p}$ represents the standard deviation of parameter $p$ across all Transformer layers at the $i_{th}$ checkpoint.

To evaluate the weight change magnitude between two checkpoints, we calculate the distance between different checkpoints' trajectories. We consider two successive trajectories, $i$ and $i+1$, and calculate the corresponding representations. We use the Fréchet distance~\cite{Frchet1906SurQP} to measure the distance between the two trajectories, $dist(p,i, i+1)$. The Fréchet distance considers the spatial and sequential variance of the curves, making it ideal for capturing differences between trajectories.

This distance is normalized using the difference in training tokens between the trajectories to account for bias and provide a relative measure of change per 1B tokens. Thus, $dist(p, i, i+1)$ is computed as follows:

$$dist(p, i, i+1) = \frac{F\_dist(std_i^p, std_{i+1}^p)}{(Token_{i+1} - Token_{i})},$$

where $Token_{i}$ denotes the token count for the $i_{th}$ checkpoint.


Fig.~\ref{parameter} suggests the following preliminary conclusions:

\begin{itemize}
\item{\textbf{Data Adjustments and Convergence.}} Despite variations in data stages, the convergence patterns of different weight matrices exhibit remarkable stability. This suggests that the proposed training methodology is robust to changes in data distribution and that data adjustments have a relatively minor impact on the overall convergence behavior of the model.

\item{\textbf{Training Efficiency.}} The convergence curves demonstrate a clear trend of decreasing mean distance between weights over time, indicating that the models converge rapidly. This is particularly evident in the later stages of training, where the curves flatten out, suggesting that the models have reached a stable state. The accelerated convergence rate can be attributed to the optimized training algorithm and hardware acceleration, leading to significant improvements in training efficiency.

\item{\textbf{Weight Matrix Behavior.}} The weight matrices associated with self-attention, MLP, and normalization layers exhibit distinct convergence patterns. While the self-attention layers converge relatively quickly, the MLP layers often require more training iterations to stabilize. This observation aligns with previous findings in the literature and highlights the complex interactions between different components of the model.
\end{itemize}

\subsection{Data Management Unit (DMU)}
Our pretraining dataset primarily consists of a blend of internet content~(Web), encyclopedic information~(Wiki), electronic books~(textbook), literary works~(paper), knowledge-intensive data~(Knowledge) from academically supervised data sets or some domain-specific knowledge and coding materials. We have explored various combinations of these diverse data sources, detailing the outcomes of these experiments to offer empirical insights and key learnings. Our goal is to offer valuable perspectives and serve as a source of inspiration for future research in this domain. The appendix.~\ref{recipe} contains an in-depth description of the experiments conducted to assess the effects of different data mixture ratios.

\textbf{Batched knowledge datasets(K6-K65)} By the experiences and lessons learned from the preliminary attempts, the final pretraining dataset is made up of three phases. The detailed data sources and processing in these datasets are shown in Table~\ref{tab:source}.

\begin{table*}[]
\caption{Data sources and processing in dataset K6-K65, \faDatabase: switching to new batches with the same data source, \faFilter: filtering data with a language model, \faPieChart: sampling data with a target data distribution, \faCloudDownload: introducing new datasets with the same type, for example, new academic datasets}
\small
\label{tab:source}
    \centering
\begin{tabular}
{|c|p{2cm}|p{1.8cm}|p{1.8cm}|p{1.8cm}|p{1.8cm}|p{1.8cm}|}
\hline
  Domain     & K6                                   & K61                          & K62                          & K63                                      & K64                                      & K65                                      \\ \hline
Web    & Wudaocorpora, Falcon refineweb       & \faDatabase+\faFilter        & \faDatabase+\faFilter        & \faDatabase+\faFilter+\faPieChart        & \faDatabase+\faFilter+\faPieChart        & \faDatabase+\faFilter+\faPieChart        \\ \hline
Wiki   & Chinese Wiki, Pile wiki              & \faDatabase+\faFilter        & \faDatabase+\faFilter        & \faDatabase+\faFilter+\faPieChart        & \faDatabase+\faFilter+\faPieChart        & \faDatabase+\faFilter+\faPieChart        \\ \hline
Paper  & Chinese Paper, Redpajama arxiv       & \faDatabase+\faFilter        & \faDatabase+\faFilter        & \faDatabase+\faFilter+\faPieChart        & \faDatabase+\faFilter+\faPieChart        & \faDatabase+\faFilter+\faPieChart        \\ \hline
Textbook & Redpajama book                     & Textbook, Teaching assistants, Masterpieces, \faFilter   & \faDatabase+\faFilter                    & \faDatabase+\faFilter+\faPieChart        & \faDatabase+\faFilter+\faPieChart        & \faDatabase+\faFilter+\faPieChart        \\ \hline
Code   & Starcoder Data                        & \faDatabase                  & \faDatabase+\faPieChart      & \faDatabase+\faPieChart                  & \faDatabase+\faPieChart                  & Tiny-codes, CodeExercises                \\ \hline
Knowledge & Chinese QA forum, Pile stackexchange & Flan 2022, COIG-PC, Academic Datasets & Instruction tuninig datasets, \faDatabase+\faCloudDownload & \faDatabase+\faCloudDownload & QA data, \faDatabase+\faCloudDownload & \faDatabase+\faCloudDownload \\ \hline
\end{tabular}
\end{table*}

\begin{itemize}
    \item {Language modeling phase (\textbf{K6})} In the first phase, the model engages in language modeling on relatively high-quality data, allowing the model parameters to converge within a reasonably optimal range. Empirically, this data set should exhibit diversity and primarily originate from a significant source, such as web data, to ensure model convergence. Furthermore, segments of relatively high quality are chosen from homogeneous data rather than random sampling. This approach is grounded in curriculum learning, where learning begins with relatively straightforward samples before progressively introducing more challenging ones. This might help prevent the model from getting stuck in local optima and accelerate the convergence process. 
    \item {Knowledge learning phase (\textbf{K61-K63})} In the knowledge learning phase, the goal of the model is to gradually strengthen its acquisition of knowledge content, building upon the foundation established in the language learning phase. To prevent excessive fluctuations in training loss, it is necessary to progressively reduce the proportion of internet-sourced data and increase the proportion of knowledge-intensive data. Additionally, due to factors like cost, non-internet data may be acquired in batches. Consequently, multiple discrete datasets need to be constructed. We developed three datasets: K61, K62, and K63, and adjusted their proportions gradually. 
    \item {Target-oriented phase (\textbf{K64, K65})} The final stage of the model seeks to combine knowledge with various language tasks, enabling it to carry out specific tasks, similar to task-specific fine-tuning. Task-specific data (e.g. academic supervised datasets) are included. The FLAN work~\cite{longpre2023flan} inspires augmenting knowledge-oriented data. Knowledge-oriented data are converted into task-specific data (QA pairs with zero-shot, few-shot, option, and no-options templates). We developed K64 for Aquila2-34B and K65 for Aquila2-70B. It is noteworthy that the 70B model opted for a different dataset K65 stemmed from observations of the trial training. The 70B model, due to budgetary constraints, did not process a sufficient number of tokens in the initial two phases. So the drastic changes in data proportions adversely affected the performance. Consequently, a compromise solution was implemented.
\end{itemize}

\begin{figure}[t]

\centering
\includegraphics[width=0.8\linewidth]{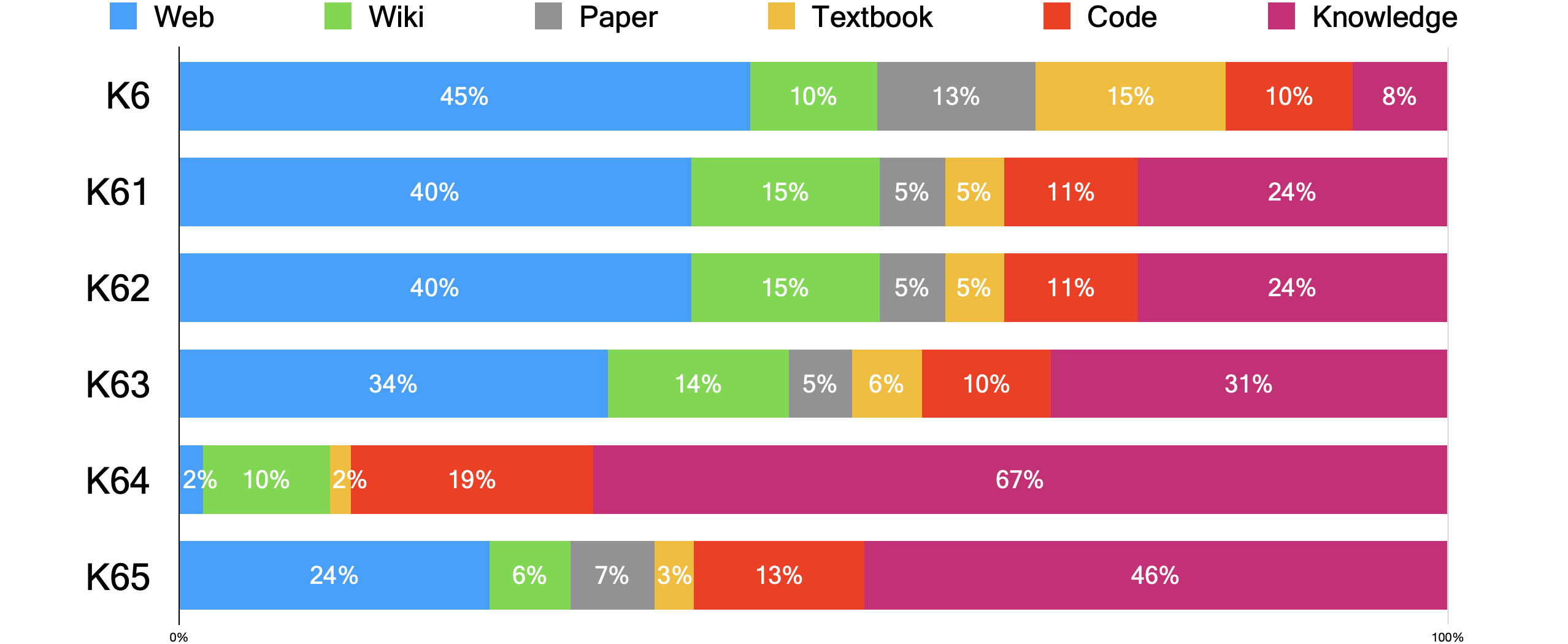}
\caption{The proportions of different domains in K6-K65\label{fig:data}}
\end{figure}

In the Aquila2-34B model, the token distributions across the three phases are as follows: 523B for the first phase, 1053B for the second phase, and 298B for the third phase. Similarly, for the Aquila2-70B model, these distributions are 203B, 719B, and 347B respectively. The varying proportions of data sourced from different origins at each stage are comprehensively depicted in Figure.~\ref{fig:data}.

The content of pretraining data can potentially reveal the fundamental causes of downstream issues such as biases, illusions, and values. Therefore, data security is of paramount importance~\cite{touvron2023llama}. While constructing the pretraining dataset, we initially performed source analysis and selection from the Chinese web corpus WuDaoCorpora Text. Only a few thousand trusted sources were chosen as candidates for the data. Subsequently, we defined types of risky data (such as content involving illegal activities, personal privacy data, harmful advice, and advertising/marketing materials) and established a manual annotation process. A security model combining classification and retrieval was trained to filter the pretraining data. Around 3\% of the data were further filtered on trusted sources.

The pretraining data was meticulously deduplicated to enhance data usage efficiency. The Aquila-34B model was trained with a total of approximately 1.8 trillion bilingual Chinese and English tokens. Compared with other open-source models such as LLaMA2-70B (2 trillion tokens), Qwen-14B (3 trillion tokens) and InternLM-20B (2.3 trillion tokens), Aquila achieved comparable results with significantly less data. Additionally, we adopted a phased approach to data preparation and training due to the lengthy data collection and training process. We adjusted the proportions of data from different domains in different batches, ensuring that the language model could better capture robust language understanding capabilities in the early stages of training. As training progressed, knowledge-based data continued to increase, allowing the model to capture richer knowledge\cite{longpre2023flan}.

Aquila2's training was conducted on the FlagScale framework~\cite{flagopen-flag-scale}. Detailed training configuration can be found in the Appendix~\ref{app:training}.

\section{Model evaluation}

In this section, we primarily exhibit the performance of our model in comparison with other models across various datasets. Initially, we juxtapose the outcomes on commonly utilized datasets, followed by a more granular analysis of the model's capabilities in different dimensions based on its specific abilities.

\subsection{Overall results}\label{evaluation metrics}

We have curated a set of open-source bilingual (Chinese-English) models, all released before December 2023, that resemble our model for comparative analysis. The primary models considered for comparison are as follows.

\begin{table}[bpht]
\setlength{\tabcolsep}{3.5pt}
\begin{tabular}{cccccccccccc}
Model           & \rotatebox{80}{Aquila2-34B} & \rotatebox{80}{Aquila2-70B} & \rotatebox{80}{Aquila2-7B}  & \rotatebox{80}{Qwen-14B} & \rotatebox{80}{LLama2-70B} & \rotatebox{80}{InternLM-20B} & \rotatebox{80}{Qwen-7B} & \rotatebox{80}{Baichuan2-13B} & \rotatebox{80}{Baichuan2-7B} & \rotatebox{80}{InternLM-7B} & \rotatebox{80}{Llama-2-7B} \\ \hline
Mean  & \textbf{72.20} & 66.51  & 58.66 & 67.71 & 67.98 & 65.99 & 61.78 & 61.93 & 57.21 & 52.48 & 46.61      \\
En Mean & \textbf{68.63} & 64.35  & 59.57 & 64.01 & 67.15 & 65.55 & 60.55 & 59.12 & 56.18 & 52.03 & 53.57      \\
Zh Mean  & \textbf{76.56} & 69.14  & 57.54 & 72.24 & 69.29 & 66.52 & 63.29 & 65.36 & 58.47 & 53.04 & 38.10      \\ \hline
WMT22 (en-zh) (5-shot)  & \textbf{61.66} & 59.93 & 59.98 & 61.04 & - & 56.90 & 60.09 & 60.50 & 55.91 & 53.25 & 36.39      \\
CLUEWSC(5-shot)   & \textbf{85.93} & 80.3  & 49.20 & 85.60 & 85.00 & 84.70 & 63.30 & 75.60 & 63.80 & 53.90 & 51.50      \\
winograde (0-shot)   & 72.85 & 70.09 & 67.50 & 67.40 & \textbf{78.00} & 75.10 & 68.59 & 70.30 & 68.43 & 68.19 & 67.09      \\
HellaSwag (10-shot)   & 82.51 & 81.35 & 71.50 & 84.00 & \textbf{87.30} & 82.10 & 78.37 & 76.10 & 73.80 & 74.35 & 78.60      \\
OpenBookQA (0-shot)  & 44.20 & 43.00 & 44.80 & 43.60 & \textbf{48.80} & 42.60 & 44.40 & 43.60 & 39.40 & 39.80 & 40.80      \\
PIQA (0-shot)     & 74.92 & 74.16 & 77.90 & 81.07 & \textbf{82.80} & 80.60 & 78.18 & 78.80 & 77.42 & 73.94 & 76.82      \\
ARC-e(0-shot)     & 79.38 & 74.66 & 71.50 & 70.41 & 81.00 & \textbf{81.10} & 73.90 & 73.90 & 72.90 & 69.99 & 53.54      \\
BUSTM(5-shot)     & \textbf{81.18} & 74.50 & 74.00 & 73.20 & 71.20 & 59.40 & 63.30 & 70.50 & 67.30 & 54.70 & 51.70      \\
BoolQ (0-shot)    & 88.84 & 86.30 & 77.60 & 86.67 & 83.70 & 82.10 & 68.01 & 79.10 & 73.00 & 43.88 & \textbf{71.13}      \\
TruthfulQA(0-shot)  & 46.99 & 38.42 & 44.80 & 49.50 & 44.80 & \textbf{51.90} & 47.85 & 39.80 & 37.55 & 39.34 & 38.80      \\
RAFT(5-shot)      & 72.50 & 69.10 & 63.20 & 68.30 & 75.20 & 75.20 & \textbf{75.40} & 71.40 & 65.50 & 61.30 & 68.60      \\
ChID(5-shot)       & \textbf{89.10} & 85.50 & 71.40 & 84.70 & 66.00 & 72.40 & 66.20 & 74.00 & 53.00 & 51.40 & 14.40      \\
EPRSTMT(5-shot)   & 90.49 & \textbf{92.80} & 89.80 & 91.20 & 89.30 & 91.20 & 91.50 & 86.60 & 88.10 & 88.10 & 59.00      \\
TNEWS(5-shot)     & \textbf{58.90} & 40.80 & 18.30 & 53.80 & 51.70 & 51.20 & 47.00 & 44.60 & 36.80 & 24.00 & 17.60      \\
OCNLI(5-shot)     & 79.92 & \textbf{81.50} & 57.00 & 55.00 & 57.60 & 62.90 & 51.00 & 43.30 & 37.10 & 36.90 & 34.20      \\
SEM-Chinese(5-shot)  & \textbf{79.50} & 74.29 & 47.63 & 77.50 & 67.20 & 72.90 & 61.08 & 64.90 & 55.83 & 51.48 & 32.45      \\
MMLU (5-shot)    & 73.74 & 53.90 & 58.10 & 65.80 & \textbf{69.50} & 61.80 & 60.16 & 56.90 & 54.60 & 51.20 & 46.90      \\
CMMLU (5-shot)   & \textbf{73.10} & 47.42 & 58.75 & 70.50 & - & 59.00 & 64.20 & 62.00 & 57.07 & 51.80 & 31.38      \\
CSL(5-shot)      & \textbf{69.19} & 66.70 & 48.80 & 52.60 & 54.60 & 51.00 & 52.90 & 49.50 & 48.40 & 48.70 & 48.50      \\
HumanEval (0-shot)  & \textbf{39.02} & 35.40 & 21.40 & 32.30 & 29.90 & 25.60 & 22.12 & 17.10 & 18.29 & 13.40 & 12.80      \\ \hline
\end{tabular}
\caption{Overall results of the Base model.}
\label{tab:res-base}
\end{table}

\begin{table*}[bpht]
\centering
\setlength{\tabcolsep}{3pt}
\begin{tabular}{cccccccccc}
\multirow{2}{*}{Model}     & Mean    & Mean  & Mean  & En Mean& Zh Mean  \\
& (Sub. \& Obj.) & (Obj.) & (Sub.) & (Obj. \& Sub.) & (Obj. \& Sub.) \\
 \hline
AquilaChat2-34B & \textbf{76.34} &  \textbf{76.38}    & \textbf{75.80}  &  \textbf{69.92}  &  \textbf{79.19}  \\

Baichuan2-13B-chat     & 63.27 &  64.14   &  58.02 &   59.09  &  65.58   \\
AquilaChat2-7B   & 63.21 &  64.02   &  58.34   &  59.38   &65.04  \\
LLaMA2-70B-chat   & 61.25 &   61.59  &  57.20 &  65.13  & 58.83  \\
InternLM-7B-chat  & 60.84 & 62.40  &  51.48  &  53.95    & 66.48  \\
Baichuan2-7B-chat & 57.99 & 58.01    & 57.87   & 55.54  & 58.28  \\
ChatGLM2-6B & 34.13 & 32.46   & 54.17  &  39.89  & 28.75   \\
\hline 
\end{tabular}

\caption{Overall results of the Chat model. "Obj." denotes objective and "Sub." denotes subjective. Detailed results can be found in the appendix.}
\label{tab:chat-res}
\end{table*}

\begin{itemize}
    \item \textbf{Baichuan2.} \cite{DBLP:journals/corr/abs-2309-10305} Baichuan2 is a series of large-scale multilingual language models, with versions containing 7 billion and 13 billion parameters. They were trained from scratch on a massive dataset of 2.6 trillion tokens. 
    \item \textbf{Qwen.} \cite{DBLP:journals/corr/abs-2309-16609} 
    Qwen is introduced as a comprehensive series of large language models (LLMs) with varying parameter counts. It includes base pre-trained models and Qwen-Chat models, which are fine-tuned with human alignment techniques for chat applications. The models in the Qwen series, such as Qwen-7B and Qwen-14B, are known for their superior performance across many downstream tasks.
    \item \textbf{LLaMA2.} \cite{DBLP:journals/corr/abs-2307-09288}
    LLaMA2 is a collection of pre-trained and fine-tuned large language models (LLMs) ranging from 7 billion to 70 billion parameters. The fine-tuned versions, called Llama 2-Chat, are optimized for dialogue use cases. It employs an auto-regressive language model architecture and is known for outperforming open-source chat models on most tested benchmarks
    \item \textbf{InternLM.} \cite{2023internlm}
    InternLM is a multilingual foundational language model with a massive 104 billion parameters, pre-trained on a large corpus consisting of 1.6 trillion tokens. The model employs a multi-phase progressive training process and is fine-tuned to align with human preferences. InternLM offers variations with different architectures and parameter sizes, such as InternLM-7B and InternLM-20B, with the latter having a deeper architecture of 60 layers compared to conventional models.
\end{itemize}

Herein, we chiefly contrast the capabilities in five aspects, namely Language, Reasoning, Understanding, Examination, and Code Generation, on prevalent datasets about these domains. The specific datasets utilized are as follows:
\begin{itemize}
    \item \textbf{Language.} WMT, CLUEWSC \cite{xu2020clue}
    \item \textbf{Reasoning.} GSM8K, winogrande \cite{sakaguchi2021winogrande}, HellaSwag \cite{zellers2019hellaswag},  OpenBookQA \cite{mihaylov2018can}, PIQA \cite{bisk2020piqa}, ARC-e \cite{clark2018think}, BUSTM \cite{xu2021fewclue}
    \item \textbf{Understanding.} BoolQ \cite{clark2019boolq}, TruthfulQA \cite{lin2021truthfulqa}, RAFT, ChID \cite{xu2021fewclue}, EPRSTMT \cite{xu2021fewclue},TNEWS \cite{xu2020clue}, OCNLI\cite{xu2020clue}, SEM-Chinese
    \item \textbf{Examination.} MMLU \cite{hendrycks2020measuring}, C-Eval \cite{huang2023ceval}, CMMLU \cite{li2023cmmlu}, CSL \cite{xu2020clue}
    \item \textbf{code genration.} HumanEval \cite{chen2021evaluating}     
\end{itemize}

To make a more fair and impartial comparison in the data sets above, we used primarily two open source third-party frameworks for evaluation: lm-evaluation-harness \cite{gao2021harness} and HELM \cite{liang2022holistic}. Specifically, Winograde, HellaSwag, OpenBookQA, PIQA, ARC-e, BoolQ, TruthfulQA, and MMLU were evaluated using the lm-evaluation-harness framework, while CLUEWSC, BUSTM, RAFT, ChID, EPRSTMT, TNEWS, OCNLI, SEM-Chinese, and CSL were assessed using the HELM framework. For the remaining datasets, evaluations were carried out using the default evaluation method recommended by the dataset.

The detailed exposition of base model results in Table \ref{tab:res-base} divulges a rich narrative concerning the comparative performance across various models on different tasks and datasets. our Aquila2-34B model demonstrates a strong performance with the highest mean score of \textbf{68.09}, indicating its robustness and generalization capabilities across a myriad of NLP tasks. In addition, Aquila2-34B scores \textbf{ 68.63} on average on English tasks and also dominates Chinese language tasks with a score of \textbf{76.56}.

The differential performance across various tasks unveils the unique strengths of each model. For instance, LLaMA2-70B shines in the Winograde, HellaSwag, and OpenBookQA tasks, suggesting its adeptness in handling reasoning and understanding-centric challenges. In contrast, the ARC-e task highlights a close competition between LLaMA2-70B and InternLM-20B, both of which hover around a score of \textbf{81}, indicating their prowess in addressing science-related queries.

In the realm of bilingual understanding, as demonstrated in the BUSTM task, Aquila2-34B emerges as a strong performer with the highest score of \textbf{81.18}. This prowess underscores the model's capability in bilingual sentence matching, which is an indispensable facet of cross-lingual NLP tasks.

The HumanEval task, albeit challenging for all models as displayed by the relatively lower scores, unveils a notable lead for Aquila2-34B with a score of \textbf{39.02}. However, this lead suggests a potential edge in mimicking human-like understanding, a frontier that remains a formidable challenge in the NLP domain.

Lastly, a close contest is observed in tasks like TNEWS, and C-Eval, indicating a balanced challenge posed by these tasks to the evaluated models. This competitive landscape underscores the importance of a thorough and diversified evaluation to glean comprehensive insights into the models' capabilities, thereby paving the way for future advancements in NLP.

\section{Conclusion and Future Work}
This research introduces Aquila2, a series of bilingual models that leverage the innovative HeuriMentor framework to optimize training efficiency and performance. HeuriMentor dynamically adjusts data distribution during training, resulting in faster convergence and improved model quality. Aquila2-34B, in particular, surpasses LLaMA-2-70B-expr and other benchmarks across 21 diverse datasets, demonstrating the efficacy of the HeuriMentor approach. Notably, Aquila2-34B maintains strong performance even under 4-bit quantization. By open-sourcing code, weights, and data, this work promotes further research in bilingual models. Future work will explore Mixture-of-Experts and data quality enhancements.

\section{Limitation}


Our evaluation process identified the inadvertent inclusion of test set data from GSM8K in the pre-training dataset. This data contamination potentially compromises the validity of Aquila2's zero-shot and few-shot results on the GSM8K benchmark.  We therefore recommend excluding these specific results from further comparisons.

To mitigate this issue, we have conducted a comprehensive re-evaluation using alternative benchmarks such as WTM22 (en-zh) and CLUEWSC. Detailed results of this re-evaluation can be found on the project's GitHub repository:\url{https://github.com/FlagAI-Open/Aquila2}.

This incident underscores the critical role of robust data validation procedures in machine learning research. Such practices ensure research integrity and the reliability of resulting AI models.





\bibliographystyle{unsrt}  
\bibliography{references}  
\section{Appendix}
\subsection{Alignment Evaluation}
The elucidation of chat model results in Table \ref{tab:chat-res} unveils a compelling narrative on the comparative efficacy of diverse chat models across various tasks and datasets. The vanguard of performance is the AquilaChat2-34B-v1.2 model, which shows a superior mean score of \textbf{76.34} (both subjective and objective), \textbf{76.38} (objective), and \textbf{75.80} (subjective). This robust performance underscores its commendable generalization capabilities across disparate NLP tasks and its proficiency in both objective and subjective evaluations.

Closer scrutiny reveals nuanced performance dynamics across different linguistic domains. In tasks associated with the English language, AquilaChat2-34B-v1.2 secures the leading position with a mean score of \textbf{69.92}, while in the Chinese language tasks, it further extends its lead, achieving a score of \textbf{79.19}. This dual linguistic prowess highlights its versatility and aptitude in handling multilingual chat-oriented tasks.

The differential performance across various tasks unveils the unique competencies of each model. For example, the BoolQ task evidences a close competition between AquilaChat2-34B and Llama-2-70B chat, both scoring in the high 80s, underscoring their ability to handle boolean questions adeptly. In contrast, in the RAFT task, Baichuan2-13B chat outweighs others with a score of \textbf{74.24}, indicating a potential strength in reasoning and factual accuracy.

In a more granular task-based analysis, the EPRSTMT task witnesses almost all models performing admirably, with scores hovering around the high 1980s and low 1990s, reflecting a shared competency in this particular task. However, the AquilaChat2-34B-v1.1 model, with a score of \textbf{91.64}, still manages to establish a lead.

Lastly, the SEM-Chinese task presents a deviation from the prevailing trend, where Baichuan2-13B-chat takes the lead with a score of \textbf{65.69}, suggesting a possible area of specialization for this model. 

The tableau of results thus furnished underscores the importance of a diverse and comprehensive evaluation regime to better understand the strengths, weaknesses, and areas of specialization inherent in each model, thereby propelling further advancements in the domain of conversational AI.

For both the base model and the supervised fine-tuned model, we prepend a special token '[CLS]' at the beginning of the input to conduct objective analysis. 
For the subjective analysis of the supervised fine-tuned model, we utilize the following template:

\textit{A chat between a curious human and an artificial intelligence assistant. The assistant gives helpful, detailed, and polite answers to the human's questions.\#\#\#Human: \textbf{\{Input\}}\#\#\#Assistant:}

All tests are performed on a single A800 GPU card. The setting of hyperparameters during testing is shown in Table~\ref{tab:generalization}:

\begin{table}[th]
\centering
\begin{tabular}{l|c|c}
\hline
Hyperparameter & Objective & Subjective \\ \hline
Top\_k & 1 & 15 \\
Top\_p & 0.95 & 0.9 \\
Temperature & 0.9 & 1 \\
Max new tokens & 1 & 512 \\
Seed & 123 & 123 \\ \hline
\end{tabular}
\caption{The hyperparameters used for objective and subjective analysis. }
\label{tab:generalization}
\end{table}

\begin{table}[th]
\centering
\begin{tabular}{l|ll|ll|l}
\hline
\multirow{2}{*}{Model}  & \multicolumn{2}{c|}{$D$}
 & \multicolumn{2}{c|}{$D_{extend}$} &\multirow{2}{*}{Avg.(\%)} \\ 
&$T_{easy}$ & $T_{hard}$ & $T_{easy}$ & $T_{hard}$  \\ \hline
LLaMA2-70B   & 46.84& 45.26 &63.42    & 52.11    & 51.99      \\
InternLM-20B &51.84 & 47.63 & 65.26    & 51.32    &  54.01   \\
Aquila2-34B  &\textbf{57.89}& 53.95 &\textbf{70.36}    & 56.05 &  59.56  \\
Aquila2-70B  &\textbf{57.89} &\textbf{58.95}  & 70.26 &  \textbf{56.58} &  \textbf{60.92}  \\  \hline
\end{tabular}
\caption{Accuracy of different models on test sets ${T_{easy}}$ and ${T_{hard}}$.}
\label{tab:generalization_evaluation}
\end{table}

\begin{table*}[t]
\small
\renewcommand{\arraystretch}{1}
\begin{center}
\begin{tabular}{c|c|c|c|c|c|c|c}
\toprule
\multirow{2}{*}{Model Name} 
& \multicolumn{2}{c|}{Inductive} & \multicolumn{2}{c|}{Deductive} & Abductive & Causal & \multirow{2}{*}{Average} \\
\cmidrule{2-7}
& bAbI-task16 & CLUTRR & bAbI-task15 & EntailmentBank & $\alpha$NLI & E-Care & \\
\midrule
Llama2-7B-Chat&        33.3&         13.3&         50.0&         80.0&         46.7&         60.0&         47.2   \\

Baichuan2-7B-Chat&        40.0&         26.7&         43.3&         73.3&         53.3&         50.0&         47.8   \\

Qwen-7B-Chat&        20.0&         10.0&         66.7&         86.7&         56.7&         56.7&         49.5  \\
Qwen-14B-Chat&        26.7&         10.0&         63.3&         86.7&         63.3&         56.7&         51.1  \\
Baichuan2-13B-Chat&        33.3&         10.0&         66.7&         80.0&         66.7&         63.3&         53.3   \\
Internlm-Chat-20B&        46.7&         13.3&         43.3&         80.0&         70.0&         70.0&         53.9  \\
GPT3.5-turbo&	36.7& 	3.3& 	93.3& 	86.7& 	53.3& 	50.0& 	53.9 \\
GPT3.5-text-davinci-002&        46.7&         6.7&         86.7&         83.3&         63.3&         46.7&         55.6  \\
Llama-2-70B-Chat&        63.3&         20.0&         53.3&         80.0&         66.7&         60.0&         57.2 \\
GPT4&        93.3&         36.7&         100&         90.0&         83.3&         83.3&         81.1  \\ 
\midrule

AquilaChat2-7B & 63.0 &	13.0 &	40.0 &	60.0 &	53.0 &	67.0 &	49.0  \\


AquilaChat2-34B& 73.3	& 30.0 & 86.7	& 73.3 & 80.0 & 76.7 & 70.0 \\

AquilaChat2-70B& 86.7	& 26.7 & 90.0	& 83.3 & 86.7 & 76.7 & 75.0 \\

\bottomrule
\end{tabular}
\caption{The evaluation results of our models(Aquila2-7B, Aquila2-34B and Aquila2-70B) on the Integrated Reasoning Dataset.}
\label{tb:reasoning}
\end{center}
\end{table*}

\subsection{Generalization Evaluation}

To evaluate the generalization performance of the models, we design a series of experiments on Aquila2-34B, Aquila2-70B, LLaMA2-70B and InternLM20B. First, we use a generic instruction tuning dataset ${D}$ to train the models and evaluate them on test set ${T_{easy}}$ across various capabilities such as language, reasoning and computation. At the same time, we create a training dataset tailored to the task types in ${T_{easy}}$. This data set is then fused with $D$ to form the combined training set ${D_{extend}}$ to evaluate the learning capacity in the domain. To further evaluate the generalization capability, we retain dimensions such as language, reasoning, and computation in ${T_{easy}}$ and adjust task difficulty to create a new test set ${T_{hard}}$. For example, the mathematical task type on ${T_{easy}}$ is to solve a linear equation of one variable, while on ${T_{hard}}$ will be adjusted to solve a linear equation of two variables.

The models are trained separately on training sets ${D_{easy}}$ and ${D_{hard}}$, and than test on ${T_{easy}}$ and ${T_{hard}}$. The results are shown in Table \ref{tab:generalization_evaluation}. Through these experiments, we could evaluate the generalizability of the models across different training data and test data of varying difficulty. 
For example, experiments in ${D}$ with ${T_{easy}}$ and ${T_{hard}}$ allow us to analyze the ability of the models to generalize from out-of-domain data to in-domain tasks. Experiments on ${D_{extend}}$ with ${T_{easy}}$ and ${T_{hard}}$ provide insight into the learning capability of the models in the domain when answering similar questions or solving more difficult questions after training on simpler ones. As shown in Table \ref{tab:generalization_evaluation}, Aquila2-34B has a 7.57\% higher average accuracy and Aquila2-70B has a 8.93\% higher average accuracy than LLaMA2-70B. Moreover, Aquila2-70B shows the highest average accuracy in the experiments.

\subsection{Reasoning Capacity Evaluation}

The proficiency of large-scale language models(LLMs) in natural language reasoning serves as a pivotal capability for the realization of Artificial General Intelligence (AGI). Logical reasoning, causal reasoning, and commonsense reasoning are three popular categories in natural language reasoning evaluation. Notably, the evaluation of commonsense reasoning primarily assesses the model's capacity for knowledge retention rather than its pure reasoning prowess. Consequently, our comprehensive assessment focused extensively on logical reasoning and causal reasoning, encompassing inductive, deductive, and abductive reasoning, including six test datasets: bAbI-task16~\cite{weston2015towards}, CLUTRR~\cite{sinha2019clutrr}, bAbI-task15~\cite{weston2015towards}, EntailmentBank~\cite{dalvi2021explaining}, $\alpha$NLI~\cite{bhagavatula2019abductive} and E-Care~\cite{du2022care}. We name the meticulously curated test data from~\cite{espejel2023gpt35} as "Integrated Reasoning Dataset(IRD)". Using the IRD data set, we subjected our AquilaChat2-34B model to stringent subjective evaluations, meticulously examining its performance against its leading industry counterparts in five publicly available tasks. Remarkably, our AquilaChat2-34B and AquilaChat2-70B demonstrate exceptional proficiency, positioning themselves as a close runner-up to GPT4.

This notable improvement in reasoning ability can be attributed to two key factors: 1. Enhanced model capacity: Models with larger parameter quantities exhibit greater potential in reasoning tasks, showcasing the significance of scale in achieving superior performance. 2. Augmented training data: The incorporation of an increased proportion of Code training data has proven instrumental in enhancing the model's reasoning capabilities. This enriched dataset contributes to a deeper understanding of complex logical structures and strengthens the model's reasoning prowess.

\begin{table}
    \centering
    \small
    \renewcommand{\arraystretch}{1}
    \setlength{\tabcolsep}{2.5pt}
    \begin{tabular}{l|ccccc}
    \hline
         & {GQA} & {POPE} & {VQAv2}  &{MM-Vet} \\
         \hline
         MiniGPT-4\cite{zhu2023minigpt4} &30.80 &- &- & 22.10 \\
         Otter\cite{li2023otter} &38.10 &- &- & 24.60 \\
         InstructBLIP\cite{dai2023instructblip} &49.20 &- &- &26.20 \\
         LLava-1.5-7b\cite{liu2023improved} & \textbf{61.90} &86.89 &\textbf{78.50} & \textbf{30.50}  \\ 
         \hline
         Aquila2VL-7B  & 61.28 &\textbf{87.05} &76.88 & 29.40 \\ 
    \hline
    \end{tabular}
    \caption{The accuracy of comparing Aquila2VL-7B with other models on different datasets.}
    \label{tab:overall_results_2}
\end{table}

\begin{table}
    \centering
    \renewcommand{\arraystretch}{1}
    \setlength{\tabcolsep}{2.5pt}
    \begin{tabular}{l|cc|ccc}
    \hline
         & {Val} & {Test} & {MCQ}  &{MRQ} &{FBQ} \\
         \hline
         
         LLava-1.5-13b & 11.36 &11.96&12.29 & 00.79 & 12.32 \\
         GPT-4V & 30.19 &30.91 &29.70 & 22.44 & \textbf{33.21} \\ 
         \hline
         Aquila2VL-34B  & \textbf{41.37} &\textbf{40.79} &\textbf{50.15} & \textbf{38.98} & 26.35 \\ 
    \hline
    \end{tabular}
    \caption{The accuracy of comparing Aquila2VL-34B with other models on different question types. MCQ means multiple-choice question, MRQ means multiple-response question and FBQ means fill-in-the-blank question.}
    \label{tab:overall_results_1}
\end{table}

\subsection{Multi-modal capability adaptation}
Broadening the scope to encompass multimodal tasks emerges as a significant application trajectory for extensive language models \cite{bai2023qwen,liu2023improved,wang2023cogvlm}. To verify the ability of the Aquila2 to acclimate to downstream tasks, we follow the experimental setting reported in LLaVA \cite{liu2023improved} to train multi-modal chat models, namely Aquila2VL-7B and Aquila2VL-34B. To evaluate the multimodal understanding ability of Aquila2VL, we evaluate GQA\cite{2019GQA}, POPE\cite{li2023evaluating}, VQAv2\cite{Goyal2019Making} and MM-Vet\cite{yu2023mmvet}. As shown in Table \ref{tab:overall_results_2}, Aquila2VL-7B achieves comparable results to LLava-1.5-7b, even outperforming POPE, achieving an accuracy of 78.50\%. To further explore the multimodal ability of Aquila2VL in Chinese, we evaluate CMMU\cite{he2024cmmu}, a benchmark for multimodal and multi-type question understanding and reasoning in Chinese. The evaluation results are shown in Table \ref{tab:overall_results_1}.  Remarkably, Aquila2VL-34B exhibits commendable against LLaVA-1.5-13B and GPT-4V, getting an accuracy of 41.37\% and 40.79\% on the validation and testing sets, respectively.
\subsection{Tokenizer}
We randomly extracted multilingual sentences from CCMatrix~\cite{schwenk2020ccmatrix}, extracted different codes from Starcoder data~\cite{li2023starcoder}, and tested and compared the average tokenized sequence length of our tokenizer and other tokenizers. The results are shown in Tab.\ref{multi} and \ref{code}, respectively.
\begin{figure}[h!]
    \centering
    \includegraphics[width=0.6\columnwidth]{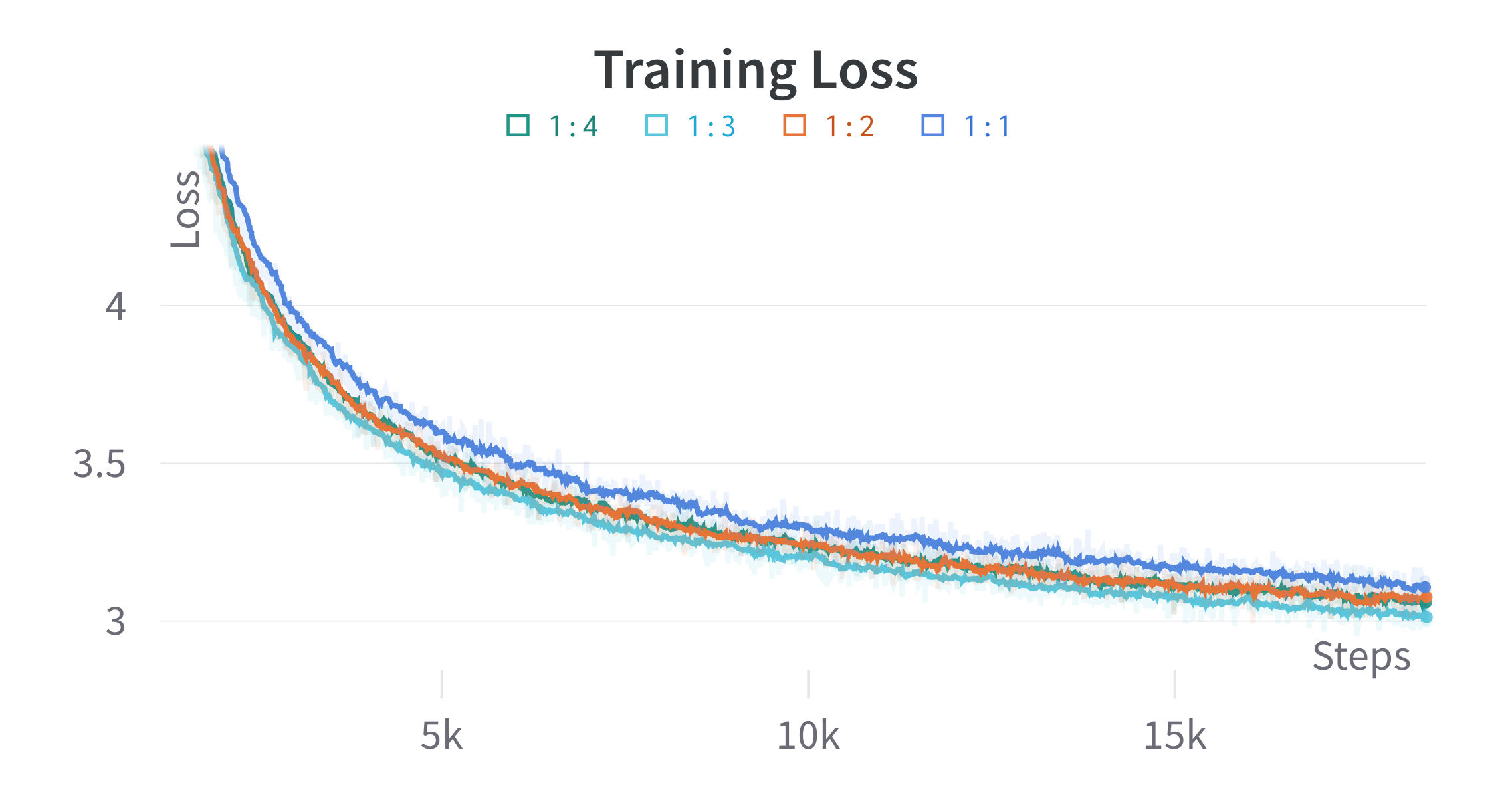}
    \caption{The training loss of 1.3B model with different Chinese and English data ratios (zh:en). }
    \label{fig:lang_ratio}
\end{figure}
\subsection{Data Recipe for Languages}
Aquila2 is a bilingual language model designed to accommodate both Chinese and English languages. 
Before commencing formal training, assessing the influence of varying ratios of Chinese and English data on the model's training performance is necessary.
To achieve this goal, we trained different 1.3B models with different proportions of Chinese and English data and observed changes in training loss.
The results are shown in Figure~\ref{fig:lang_ratio}.
We found that generally speaking, the higher the proportion of Chinese in the training data, the greater the overall loss of model training. This may to some extent prove that the difficulty of learning Chinese corpus is greater than that of English.
In our investigation, we observed that, for our dataset, maintaining a token ratio of approximately 1:2 or 1:3 between the Chinese and English training corpora led to lower training loss. 
Consequently, during formal training, we establish and adjust the proportion of Chinese and English corpora based on both this finding and the performance metrics of downstream tasks.

\begin{table*}[]
\setlength{\tabcolsep}{1mm}
\centering
\begin{tabular}{lllllllllllllll} \hline
                & Ar   & De   & En   & Es   & Fr   & Hi   & It   & Ja   & Ko   & Pt   & Zh   & Ru    & Th    & Average \\\hline
Chatglm2(64K)   & 74.3 & 36.2 & 19.8 & 35.1 & 30.1 & 74.8 & 36.7 & 27.0 & 35.4 & 34.3 & 15.8 & 102.1 & 140.5 & 38.1    \\
Baichuan(64K)   & 74.3 & 39.6 & 20.9 & 37.6 & 32.6 & 69.6 & 39.3 & 27.4 & 34.4 & 36.7 & 17.9 & 116.9 & 137.0 & 39.1    \\
Baichuan2(125K) & 69.5 & 36.7 & 19.7 & 35.4 & 31.0 & 66.3 & 37.0 & 25.4 & 33.0 & 34.6 & 15.2 & 106.6 & 135.0 & 36.7    \\
Aquila(100K)    & 58.5 & 34.7 & 19.1 & 32.2 & 28.6 & 66.1 & 34.4 & 25.5 & 49.8 & 31.7 & 15.9 & 107.2 & 154.9 & 36.0    \\
Qwen(256k)      & 31.2 & 29.7 & 19.1 & 28.2 & 25.3 & 63.7 & 31.9 & 18.1 & 23.2 & 27.6 & 15.9 & 77.2  & 78.2  & 28.5    \\\hline
\end{tabular}
\caption{The average tokenized sequence length for different languages. \label{multi}}
\end{table*}

\begin{table*}[]
\setlength{\tabcolsep}{1mm}
\centering
\begin{tabular}{llllllllll}\hline
Tokenizer     & Shell & Tex   & Java  & Python & Js    & SQL   & C     & Cpp   & Average \\\hline
Ziya(40K)     & 386.8 & 758.9 & 679.0 & 708.3  & 543.0 & 510.3 & 739.7 & 878.3 & 650.5   \\
Chatglm2(64K) & 402.7 & 780.0 & 691.4 & 716.1  & 555.6 & 529.1 & 744.5 & 889.1 & 663.6   \\
Baichuan(64K) & 431.8 & 835.2 & 723.0 & 761.7  & 586.9 & 567.9 & 787.4 & 943.5 & 704.7   \\
Aquila2(100K) & 353.6 & 684.4 & 594.8 & 623.9  & 473.7 & 448.5 & 644.6 & 767.8 & 573.9   \\
Qwen(256k)    & 303.4 & 674.3 & 488.9 & 526.0  & 411.3 & 392.9 & 538.1 & 647.0 & 497.7   \\\hline
\end{tabular}
\caption{The average tokenized sequence length for different codes.\label{code}}
\end{table*}
\subsection{Case Study}
\begin{figure*}
    \centering
    \includegraphics[width=0.95\textwidth]{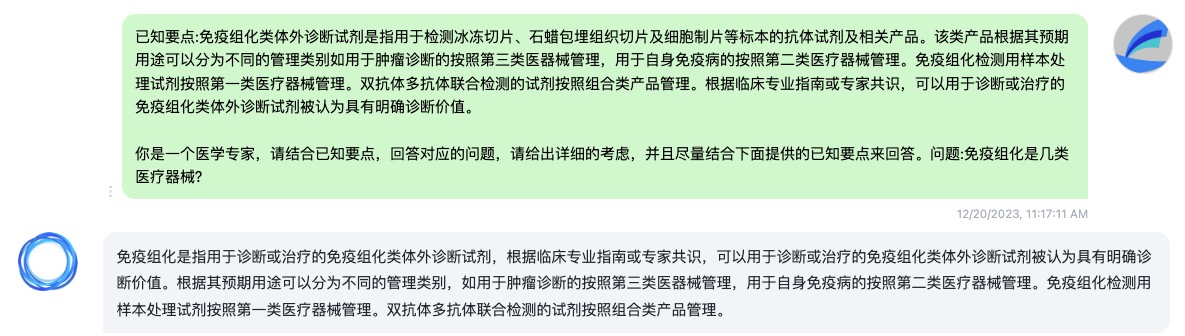}
    \caption{Question answer example.}
    \label{fig:case1}
\end{figure*}

\begin{figure*}
    \centering
    \includegraphics[width=0.95\textwidth]{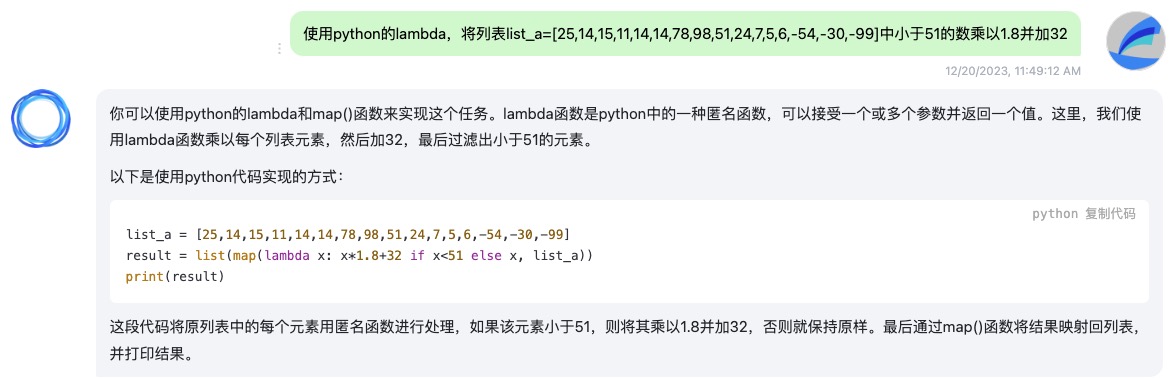}
    \caption{Code generation example.}
    \label{fig:case5}
\end{figure*}

\begin{figure*}
    \centering
    \includegraphics[width=0.95\textwidth]{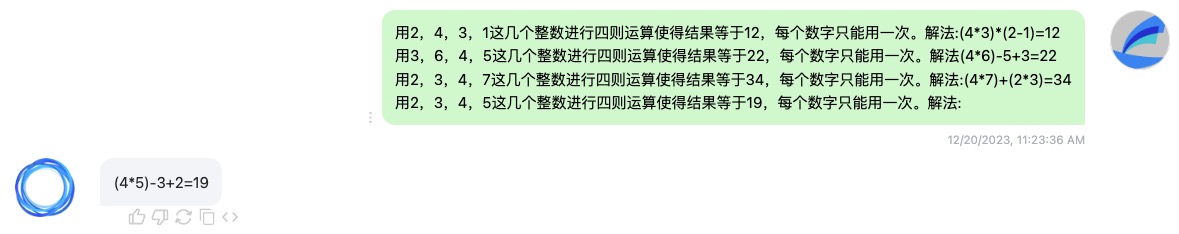}
    \caption{Mathematical reasoning example.}
    \label{fig:case2}
\end{figure*}

\begin{figure*}
    \centering
    \includegraphics[width=0.95\textwidth]{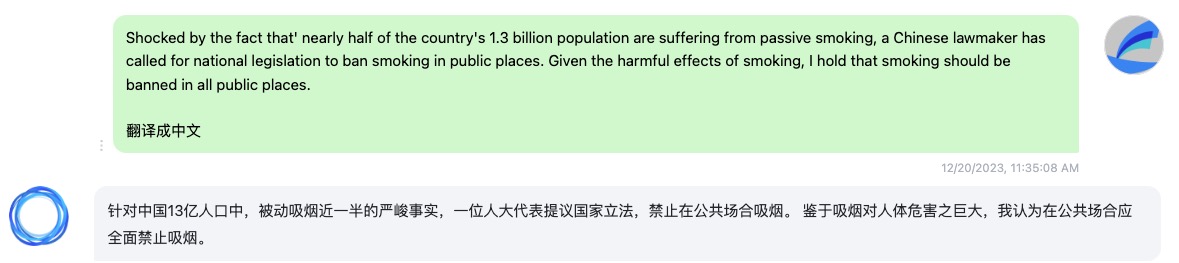}
    \caption{Translation example.}
    \label{fig:case3}
\end{figure*}

\begin{figure*}
    \centering
    \includegraphics[width=0.95\textwidth]{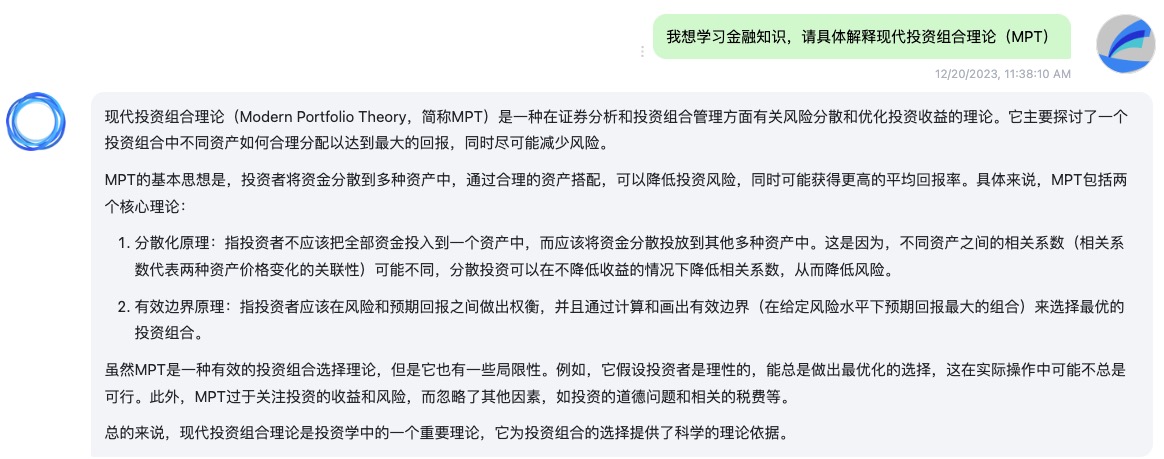}
    \caption{Knowledge explanation example.}
    \label{fig:case4}
\end{figure*}

\begin{figure*}
    \centering
    \includegraphics[width=0.95\textwidth]{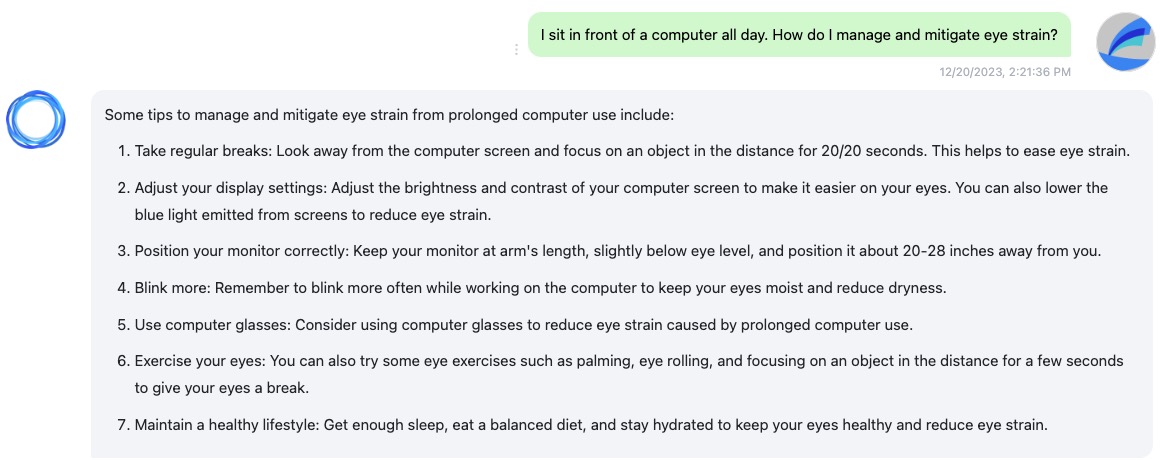}
    \caption{Advice consultation example.}
    \label{fig:case6}
\end{figure*}

\begin{figure*}
    \centering
    \includegraphics[width=0.95\textwidth]{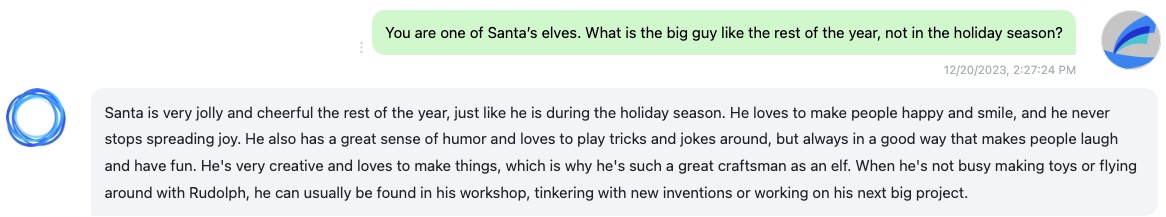}
    \caption{Role play example.}
    \label{fig:case7}
\end{figure*}

We provide some generation examples of our AquilaChat2-34B-v1.2 model, which are shown as in Figure~\ref{fig:case1},~\ref{fig:case5},~\ref{fig:case2},~\ref{fig:case3},~\ref{fig:case4},~\ref{fig:case6} and \ref{fig:case7}.

\begin{table*}
    \centering
    \begin{tabular}{lr}
        \toprule
        Configuration Key & Value \\
        \midrule
        adam\_beta1 & 0.9 \\
        adam\_beta2 & 0.95 \\
        adam\_eps & 1e-08 \\
        add\_bias\_linear & False \\
        apply\_layernorm\_rms & True \\
        apply\_query\_key\_layer\_scaling & True \\
        apply\_residual\_connection\_post\_layernorm & False \\
        attention\_dropout & 0.0 \\
        attention\_softmax\_in\_fp32 & True \\
        bf16 & True \\
        bias\_dropout\_fusion & True \\
        data\_parallel\_size  & 32 \\
        distributed\_backend & nccl \\
        embedding\_weights\_in\_fp32 & True \\
        ffn\_hidden\_size & 24576 \\
        global\_batch\_size & 1024 \\
        gradient\_accumulation\_fusion & True \\
        group\_query\_attention & True \\
        hidden\_dim\_multiplier & 1.3 \\
        hidden\_dropout & 0.0 \\
        hidden\_size & 6144 \\
        init\_method\_std & 0.0165 \\
        layernorm\_epsilon & 1e-05 \\
        layernorm\_init\_weight & 0.3 \\
        lr & 0.00015 \\
        lr\_decay\_style & cosine \\
        lr\_warmup\_samples & 500000 \\
        make\_vocab\_size\_divisible\_by & 64 \\
        max\_position\_embeddings & 4096 \\
        micro\_batch\_size & 1 \\
        min\_lr & 1.5e-05 \\
        num\_attention\_heads & 64 \\
        num\_layers & 60 \\
        num\_query\_groups & 8 \\
        optimizer & adam \\
        pipeline\_model\_parallel\_size & 4 \\
        position\_embedding\_type & rope \\
        rampup\_batch\_size & 32, 32, 2000000 \\
        rotary\_percent & 1.0 \\
        rotary\_position\_embeddings\_in\_fp32 & True \\
        save\_interval & 1000 \\
        seed & 42 \\
        seq\_length & 4096 \\
        sequence\_parallel & True \\
        split & 1 \\
        swiglu & True \\
        tensor\_model\_parallel\_size & 4\\
        tokenizer\_type & AquilaTokenizer \\
        use\_distributed\_optimizer & True \\
        use\_flash\_attn & True \\
        vocab\_size & 100008 \\
        weight\_decay & 0.1 \\
        weight\_decay\_incr\_style &constant \\
        \bottomrule
    \end{tabular}
    \caption{Full configuration of Aquila2-34B pretraining}
    \label{tab:aquila34b-config}
\end{table*}   

\begin{table*}
    \centering
    \begin{tabular}{lr}
        \toprule
        Configuration Key & Value \\
        \midrule
        adam\_beta1 & 0.9 \\
        adam\_beta2 & 0.95 \\
        adam\_eps & 1e-08 \\
        add\_bias\_linear & False \\
        apply\_layernorm\_rms & True \\
        apply\_query\_key\_layer\_scaling & True \\
        apply\_residual\_connection\_post\_layernorm & False \\
        attention\_dropout & 0.0 \\
        attention\_softmax\_in\_fp32 & True \\
        bf16 & True \\
        bias\_dropout\_fusion & True \\
        data\_parallel\_size  & 24 \\
        distributed\_backend & nccl \\
        embedding\_weights\_in\_fp32 & True \\
        ffn\_hidden\_size & 28672 \\
        global\_batch\_size & 1056 \\
        gradient\_accumulation\_fusion & True \\
        group\_query\_attention & True \\
        hidden\_dim\_multiplier & 1.3 \\
        hidden\_dropout & 0.0 \\
        hidden\_size & 8192 \\
        init\_method\_std & 0.0149 \\
        layernorm\_epsilon & 1e-05 \\
        layernorm\_init\_weight & 0.25 \\
        lr & 0.00015 \\
        lr\_decay\_style & cosine \\
        lr\_warmup\_samples & 500000 \\
        make\_vocab\_size\_divisible\_by & 64 \\
        max\_position\_embeddings & 4096 \\
        micro\_batch\_size & 2 \\
        min\_lr & 1.5e-05 \\
        num\_attention\_heads & 64 \\
        num\_layers & 80 \\
        num\_query\_groups & 8 \\
        optimizer & adam \\
        pipeline\_model\_parallel\_size & 4 \\
        position\_embedding\_type & rope \\
        rampup\_batch\_size & 48, 48, 2000000 \\
        rotary\_percent & 1.0 \\
        rotary\_position\_embeddings\_in\_fp32 & True \\
        save\_interval & 500 \\
        seed & 42 \\
        seq\_length & 4096 \\
        sequence\_parallel & True \\
        split & 1 \\
        swiglu & True \\
        tensor\_model\_parallel\_size & 8 \\
        tokenizer\_type & AquilaTokenizer \\
        use\_distributed\_optimizer & True \\
        use\_flash\_attn & True \\
        vocab\_size & 100008 \\
        weight\_decay & 0.1 \\
        weight\_decay\_incr\_style &constant \\
        \bottomrule
    \end{tabular}
    \caption{Full configuration of Aquila2-70B pretraining}
    \label{tab:aquila70b-config}
\end{table*}  

\subsection{Convergence Observation via Weights}
\label{parameter_observation}
\begin{figure*}[hbpt]
\centering
\begin{subfigure}[K6]{\includegraphics[width=0.48\linewidth]{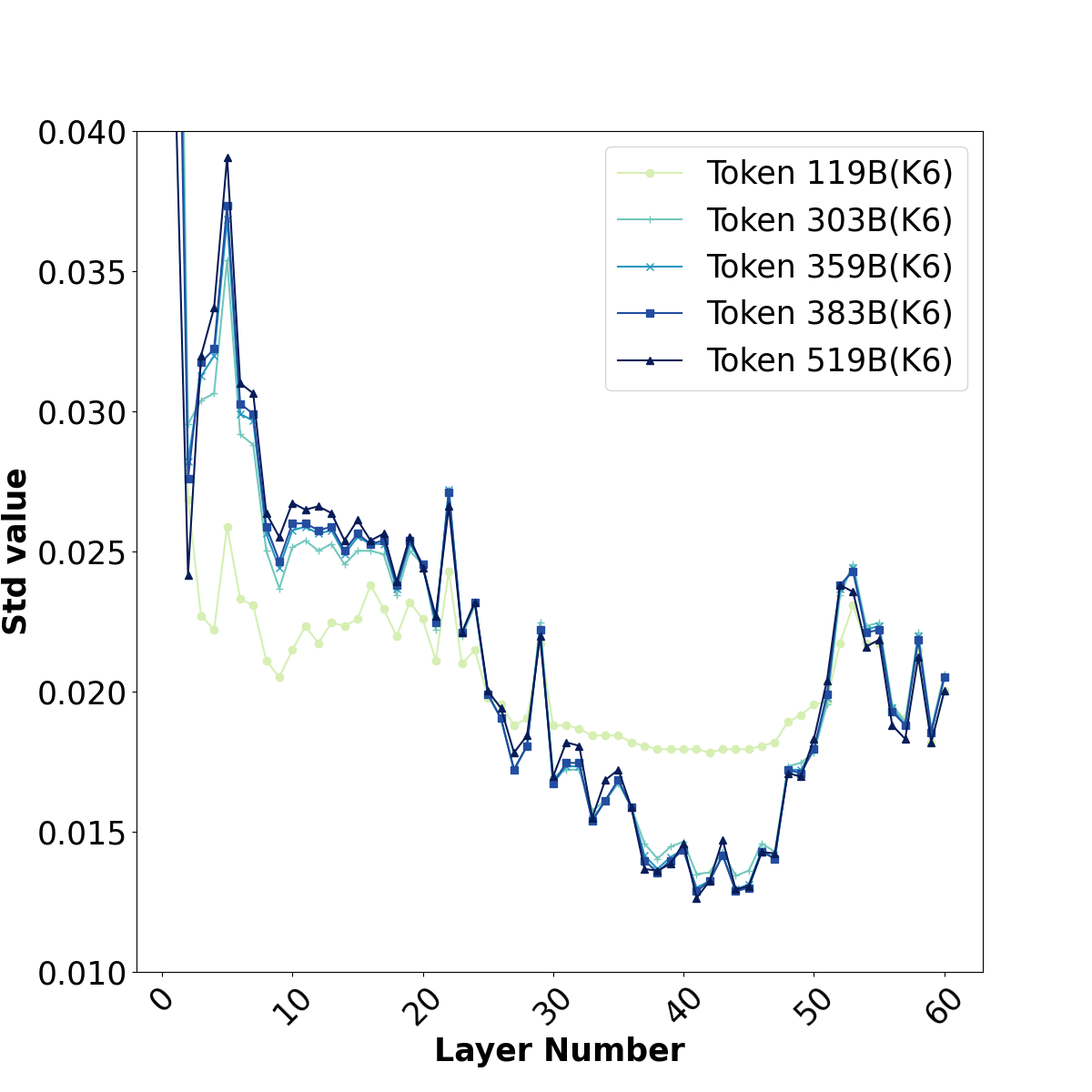}}
\end{subfigure}
\hfill
\begin{subfigure}[K61$\&$K62]{\includegraphics[width=0.48\linewidth]{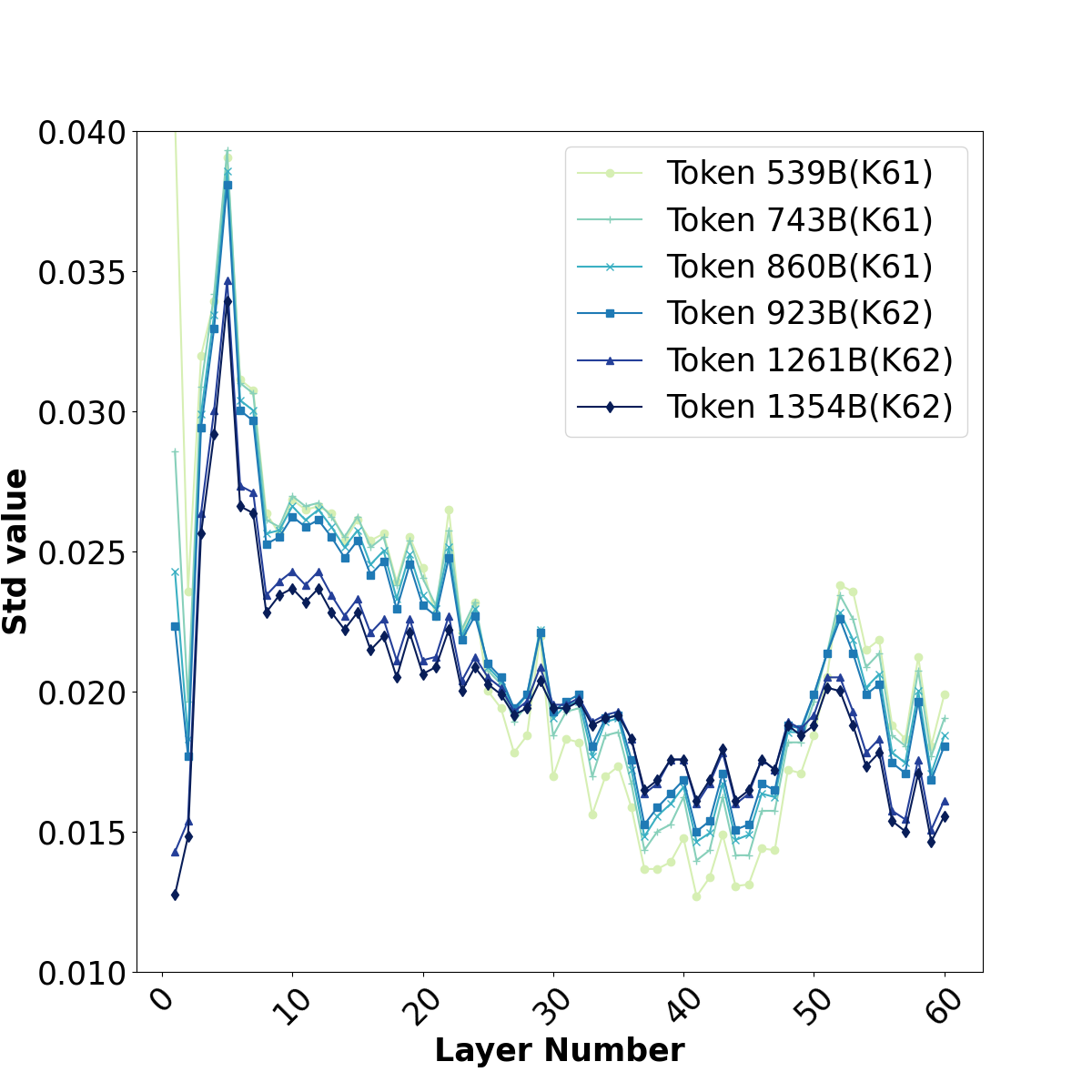}}
\end{subfigure}

\begin{subfigure}[K63]{\includegraphics[width=0.48\linewidth]{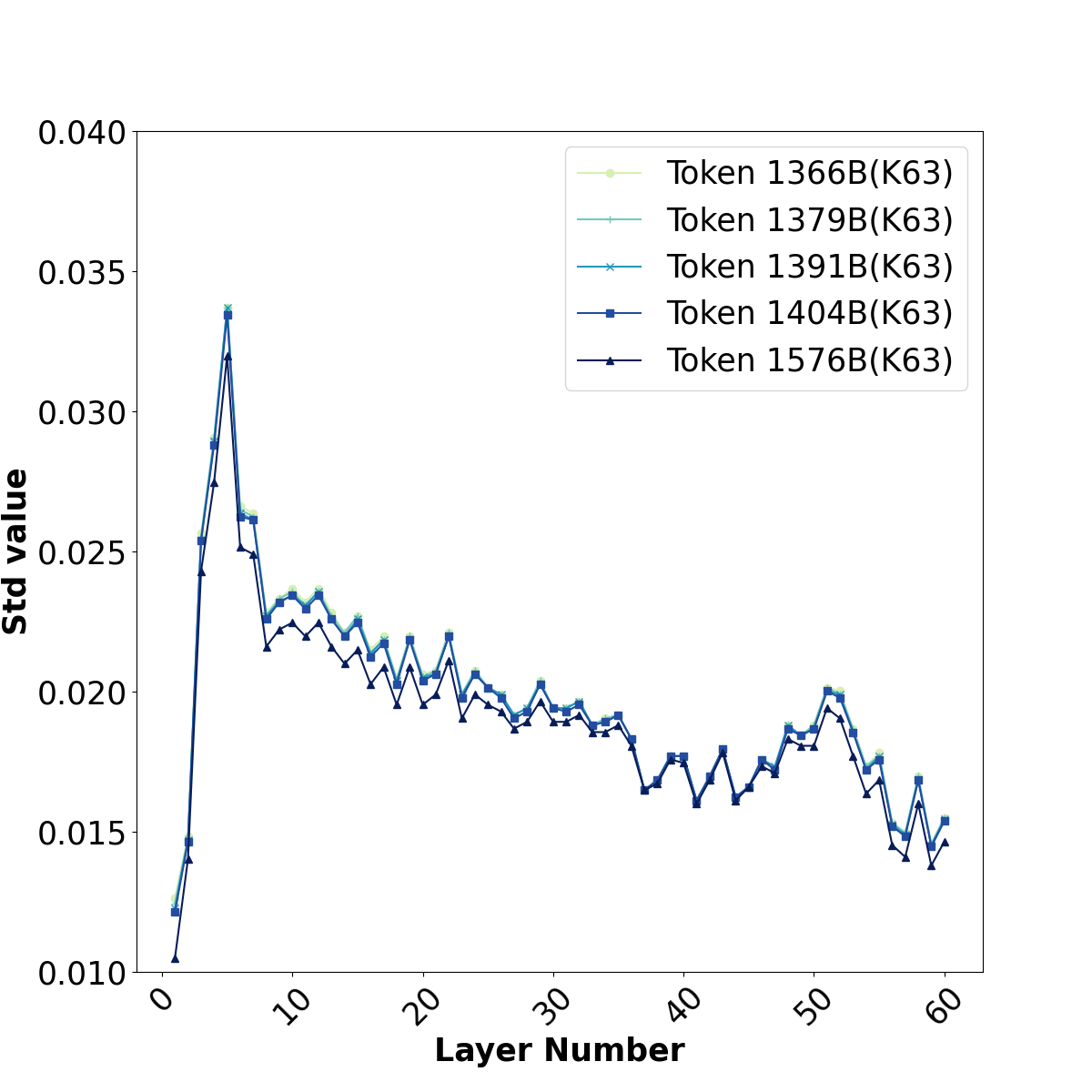}}
\end{subfigure}
\hfill
\begin{subfigure}[K64]{\includegraphics[width=0.48\linewidth]{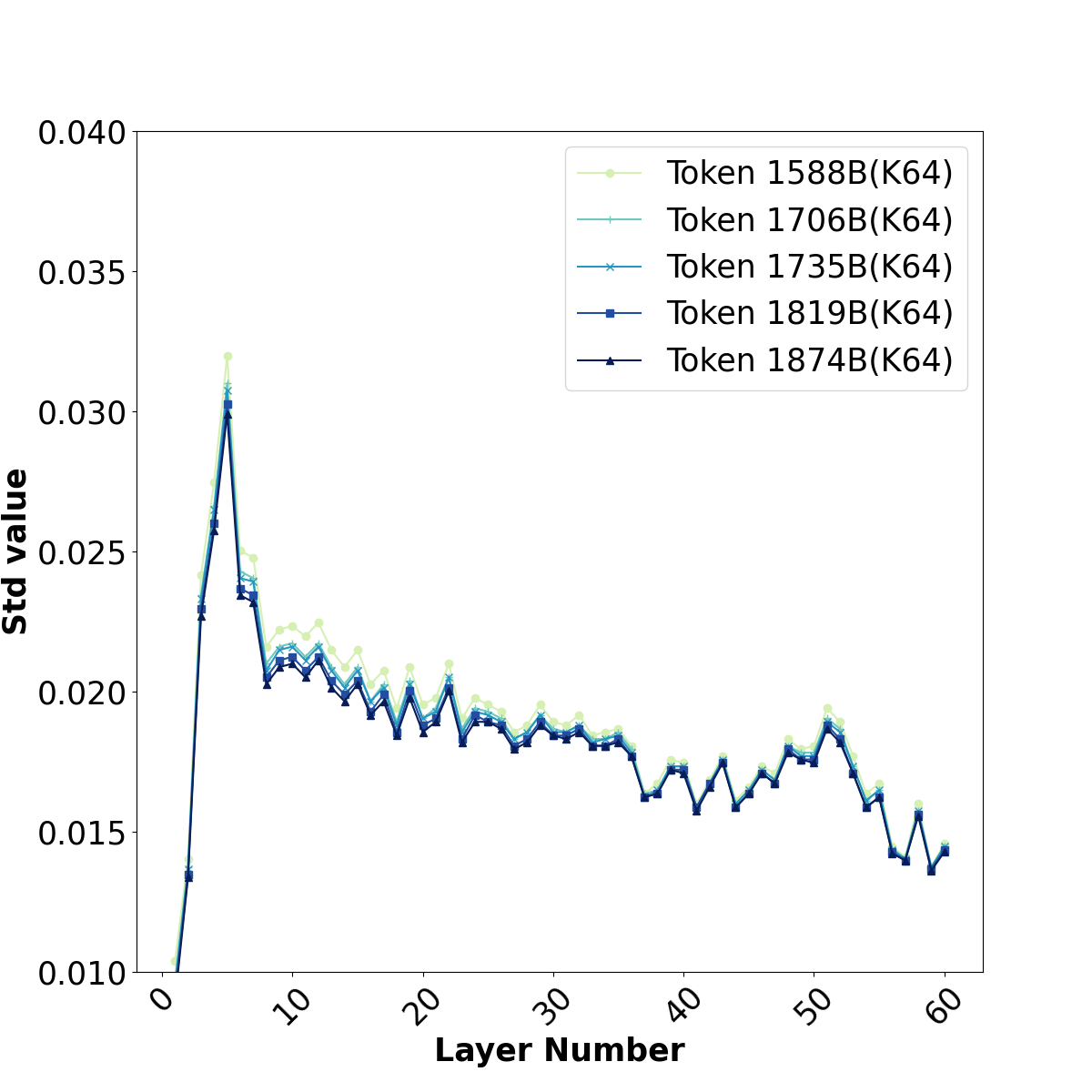}}
\end{subfigure}

\caption{Standard variance of the weight matrix $W_K$ across different layers in Aquila-34B. Different tokens in the graph represent the total amount of training data tokens used.}
\label{34B_k}
\end{figure*}

\begin{figure*}[hbpt]
\centering
\begin{subfigure}[K6]{\includegraphics[width=0.48\linewidth]{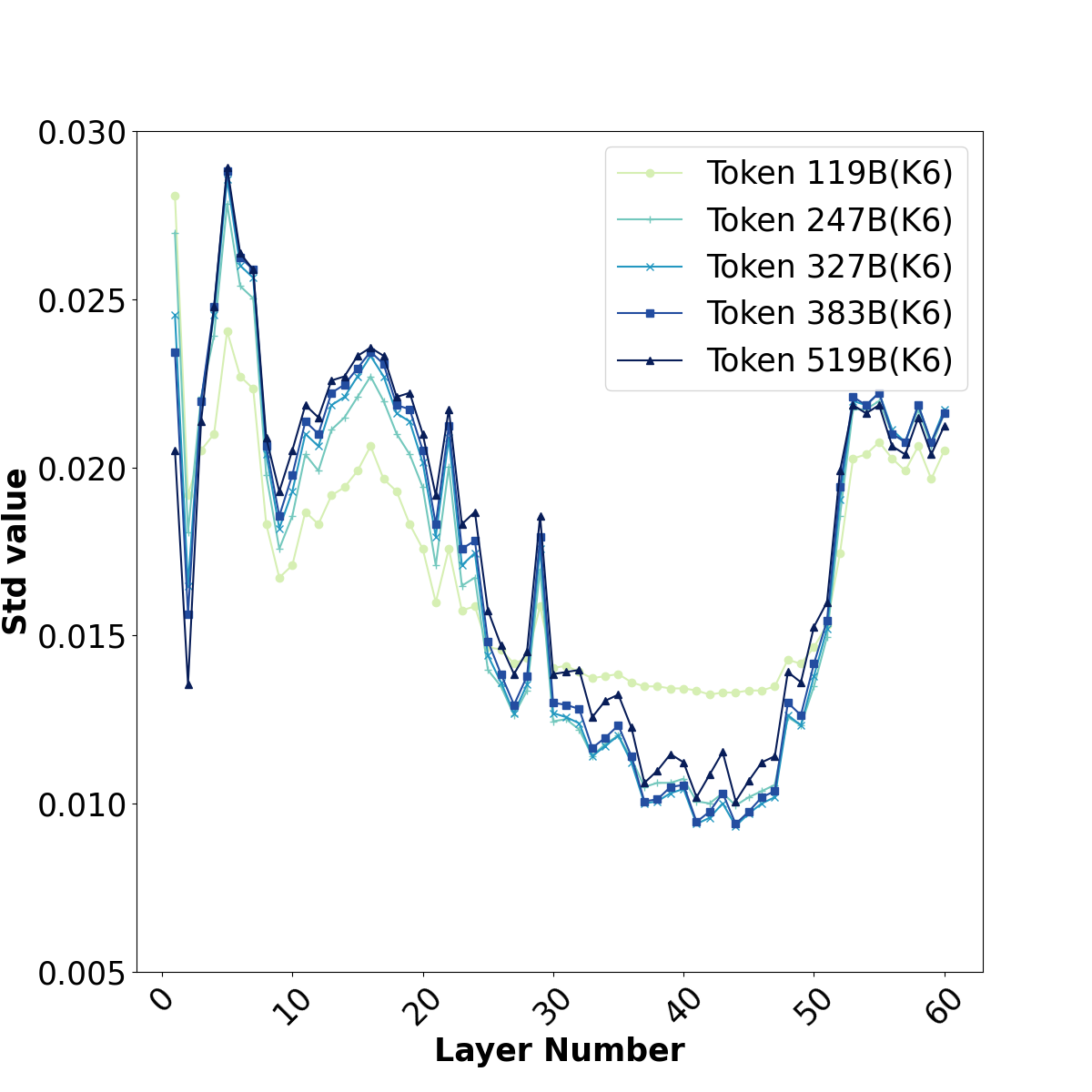}}
\end{subfigure}
\hfill
\begin{subfigure}[K61$\&$K62]{\includegraphics[width=0.48\linewidth]{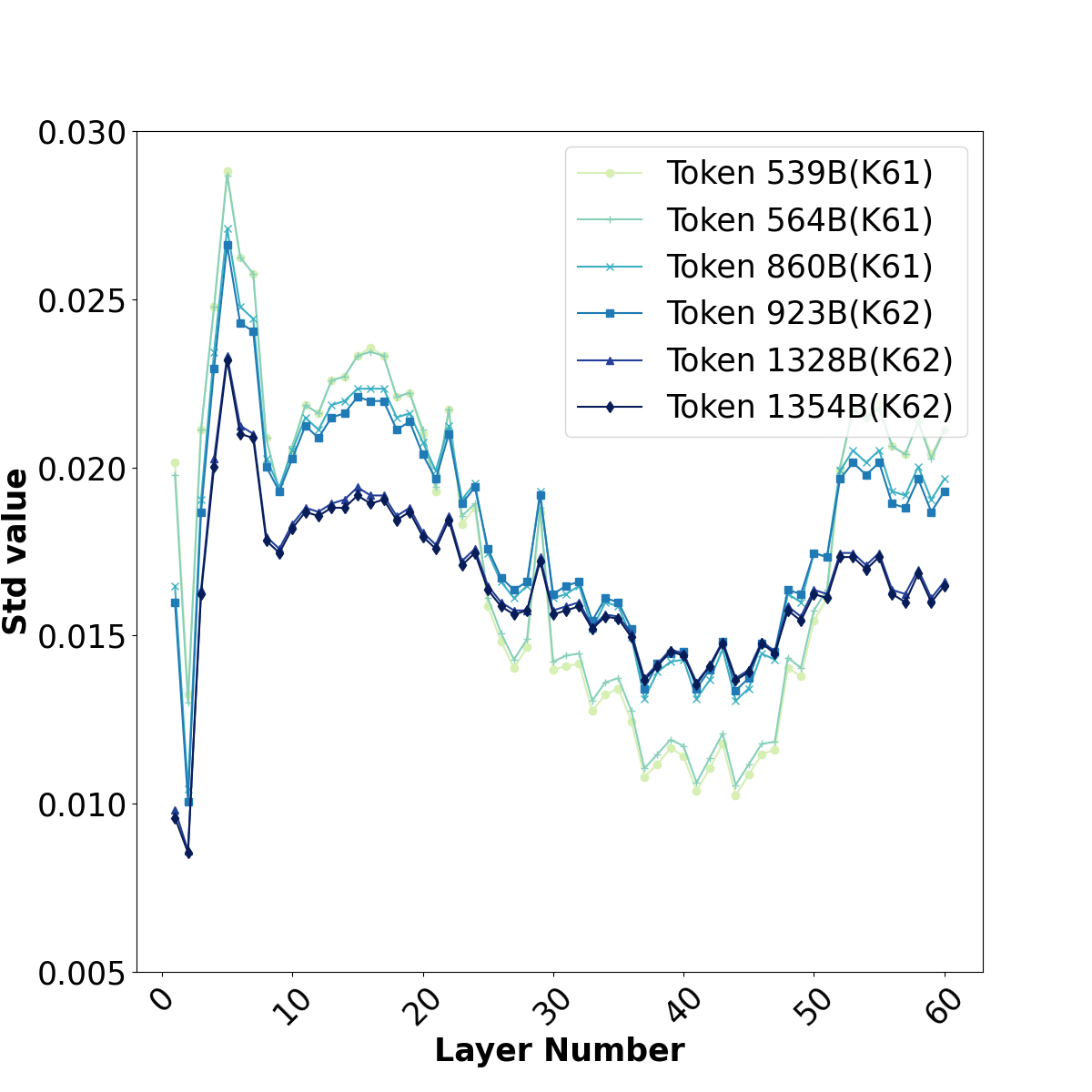}}
\end{subfigure}

\begin{subfigure}[K63]{\includegraphics[width=0.48\linewidth]{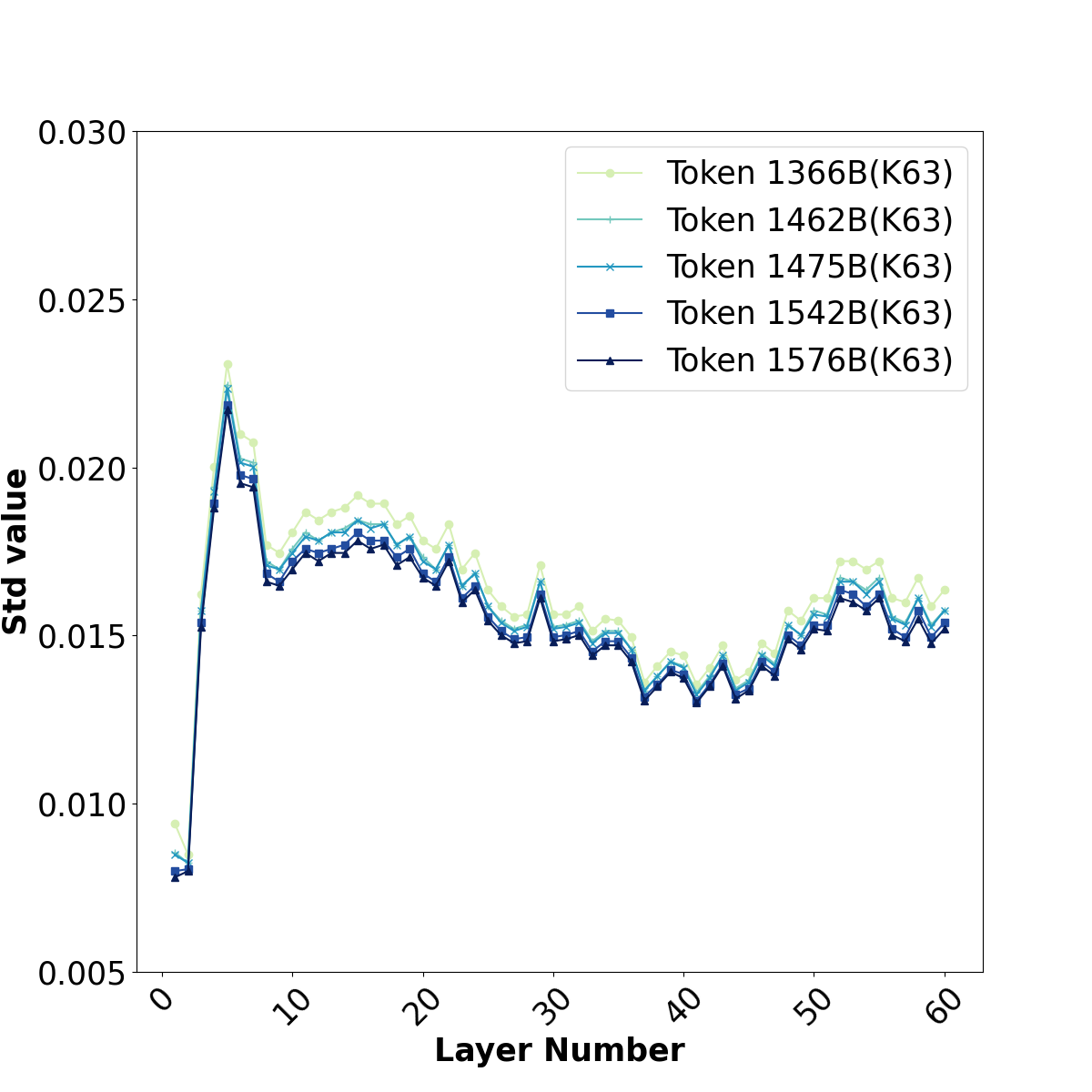}}
\end{subfigure}
\hfill
\begin{subfigure}[K64]{\includegraphics[width=0.48\linewidth]{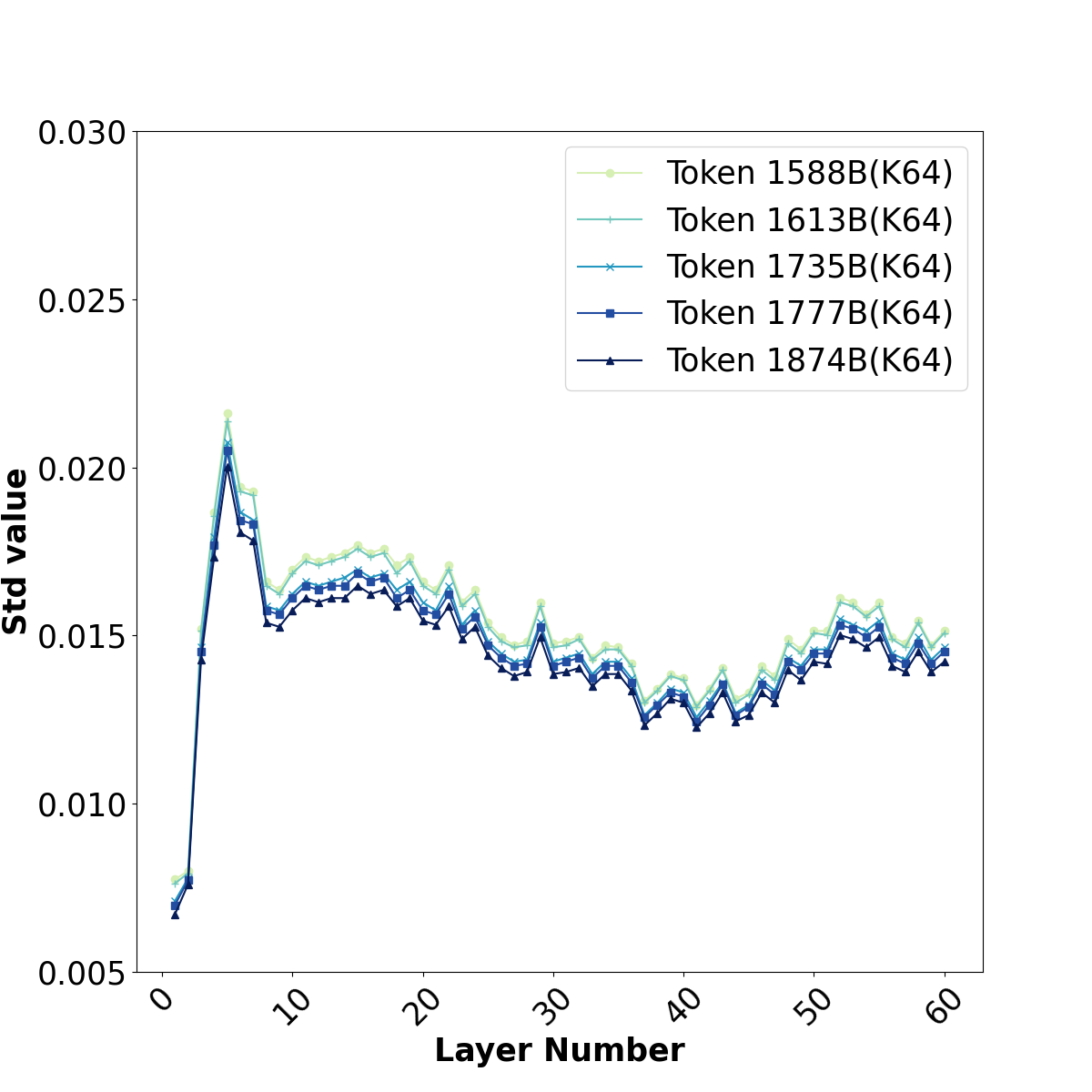}}
\end{subfigure}

\caption{Standard variance of the weight matrix $W_Q$ across different layers in Aquila-34B. Different tokens in the graph represent the total amount of training data tokens used.}
\label{34B_q}
\end{figure*}

\begin{figure}[hbpt]
\centering
\begin{subfigure}[K6]{\includegraphics[width=0.48\linewidth]{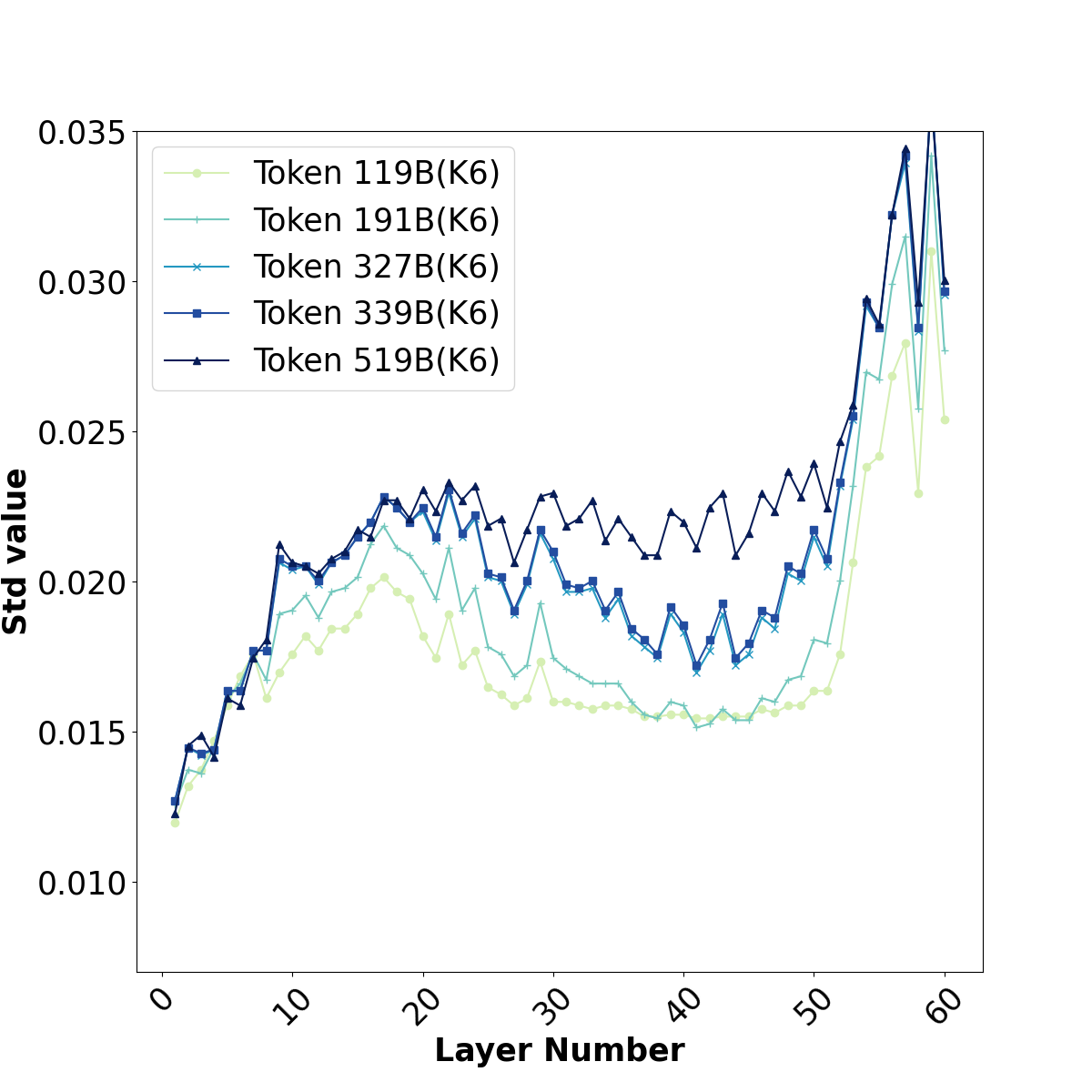}}
\end{subfigure}
\hfill
\begin{subfigure}[K61$\&$K62]{\includegraphics[width=0.48\linewidth]{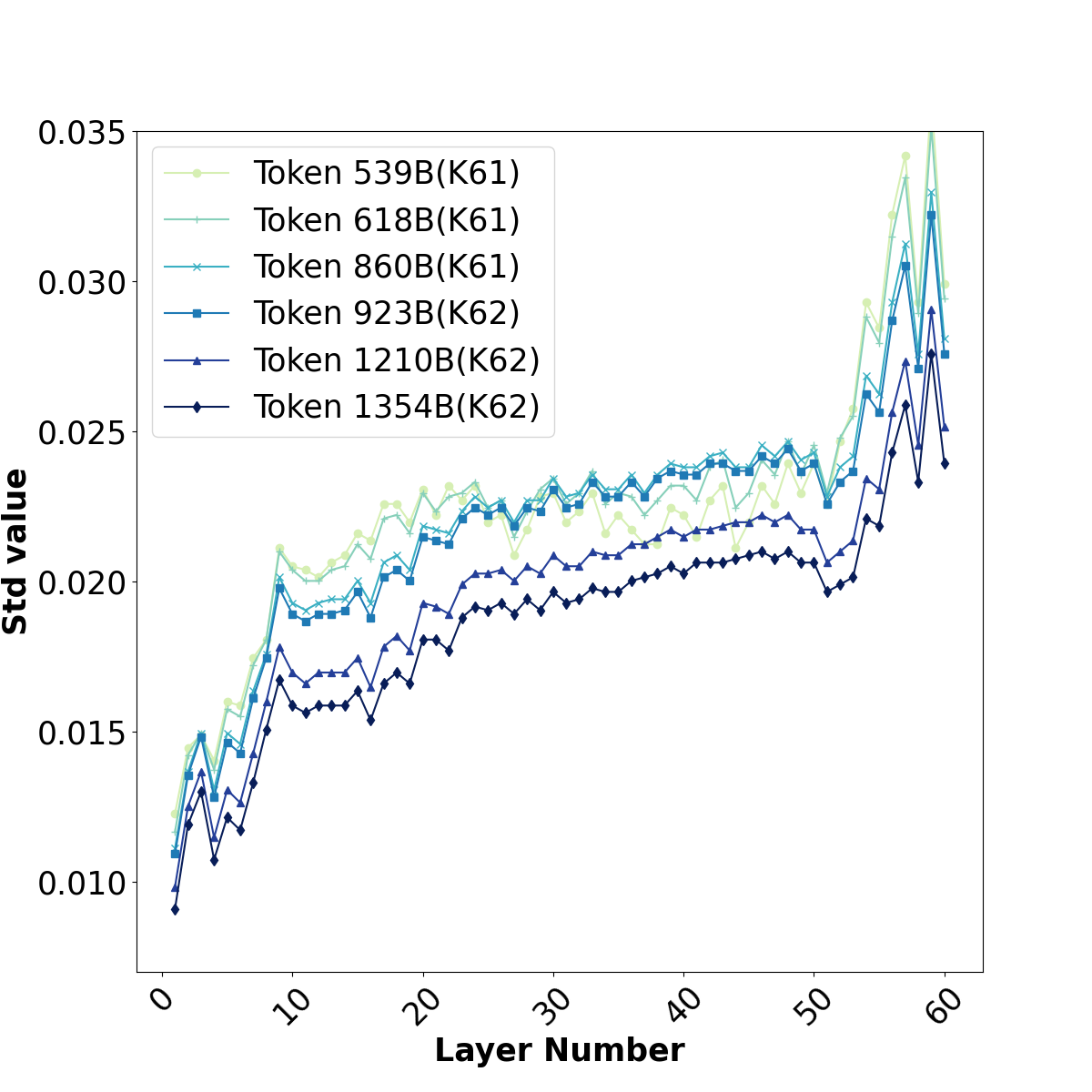}}

\end{subfigure}
\begin{subfigure}[K63]{\includegraphics[width=0.48\linewidth]{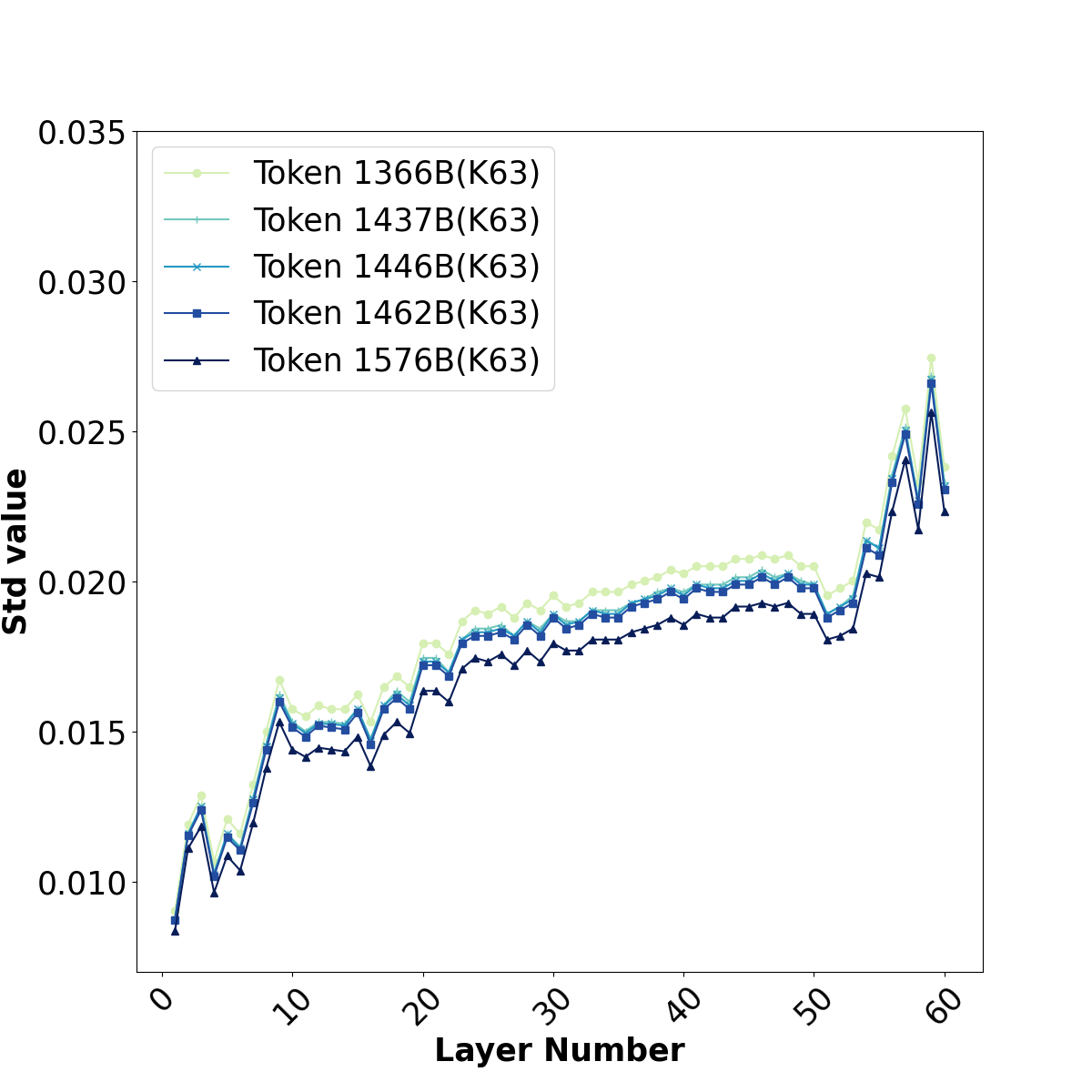}}
\end{subfigure}
\hfill
\begin{subfigure}[K64]{\includegraphics[width=0.48\linewidth]{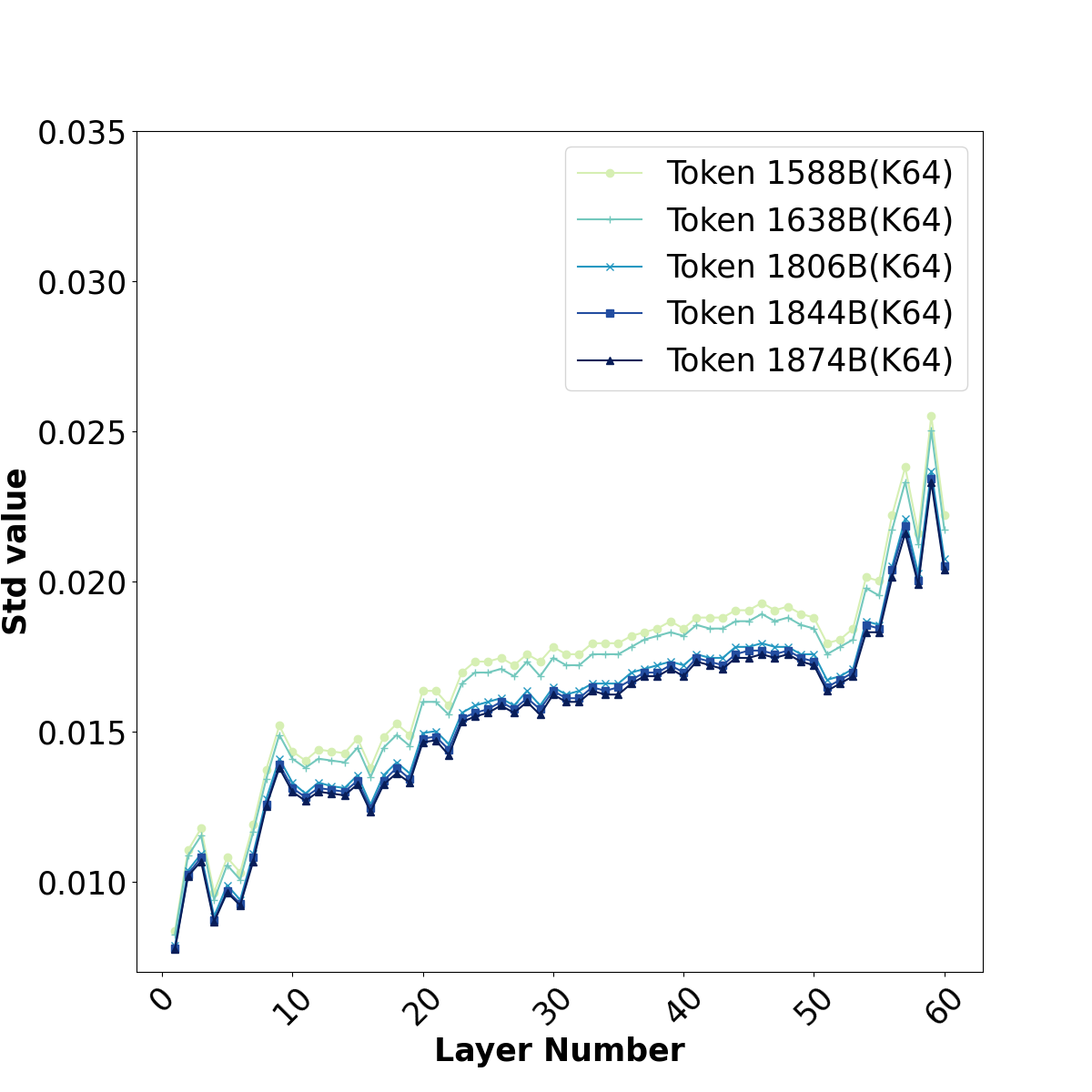}}
\end{subfigure}

\caption{Standard variance of the weight matrix $W_V$ across different layers in Aquila-34B. Different tokens in the graph represent the total amount of training data tokens used.}
\label{34B_v}
\end{figure}

\begin{figure*}[hbpt]
\centering
\begin{subfigure}[K6]{\includegraphics[width=0.48\linewidth]{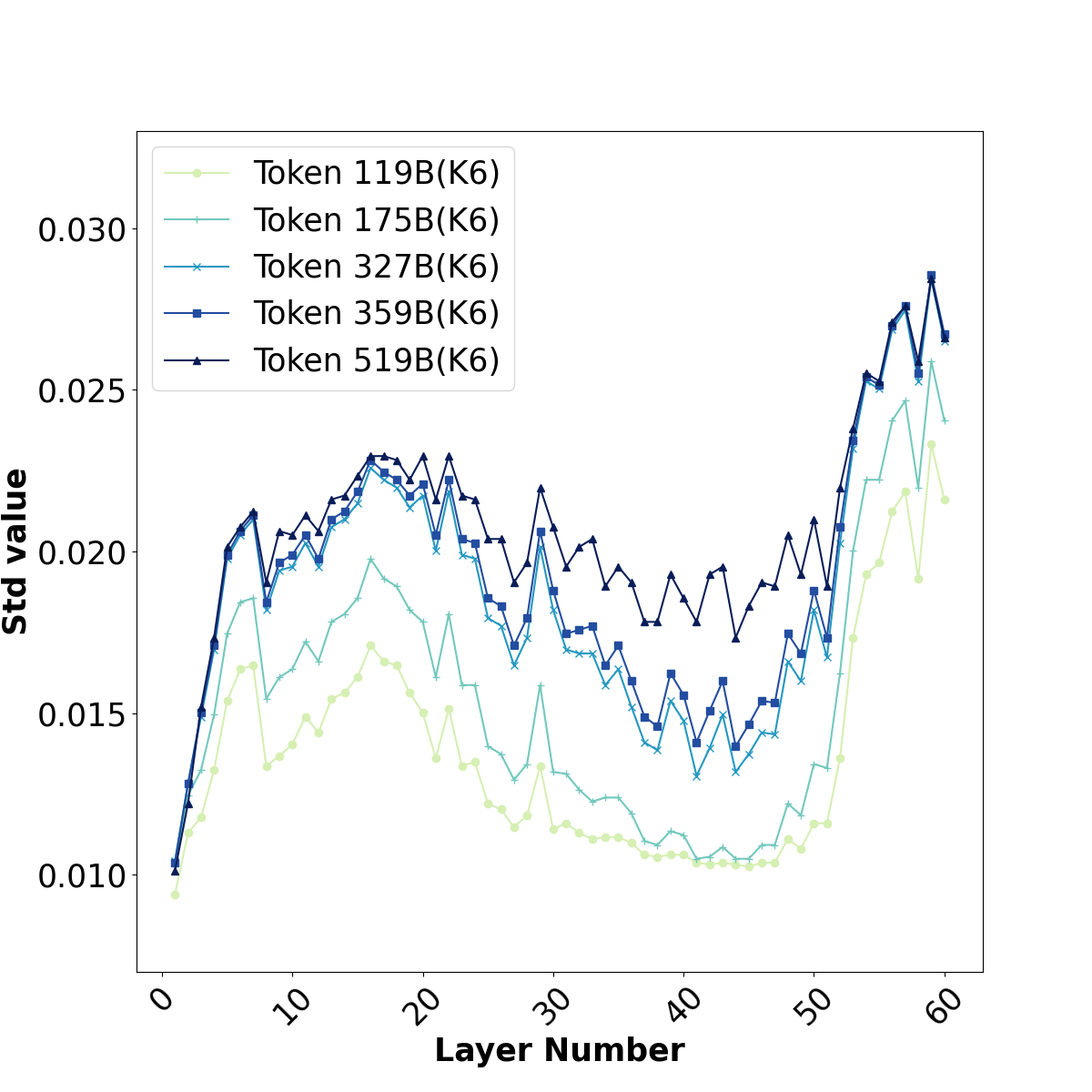}}
\end{subfigure}
\hfill
\begin{subfigure}[K61$\&$K62]{\includegraphics[width=0.48\linewidth]{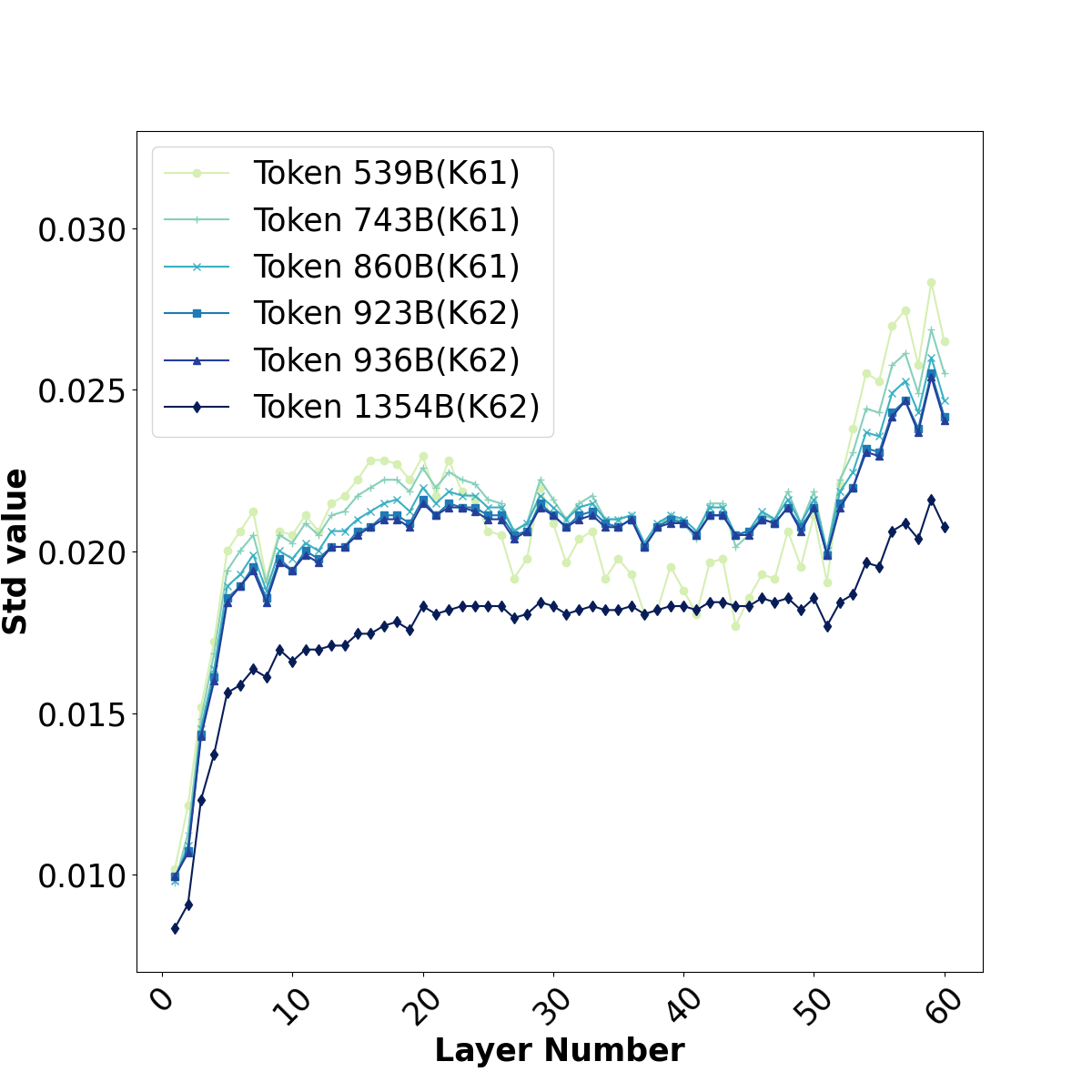}}
\end{subfigure}
\begin{subfigure}[K63]{\includegraphics[width=0.48\linewidth]{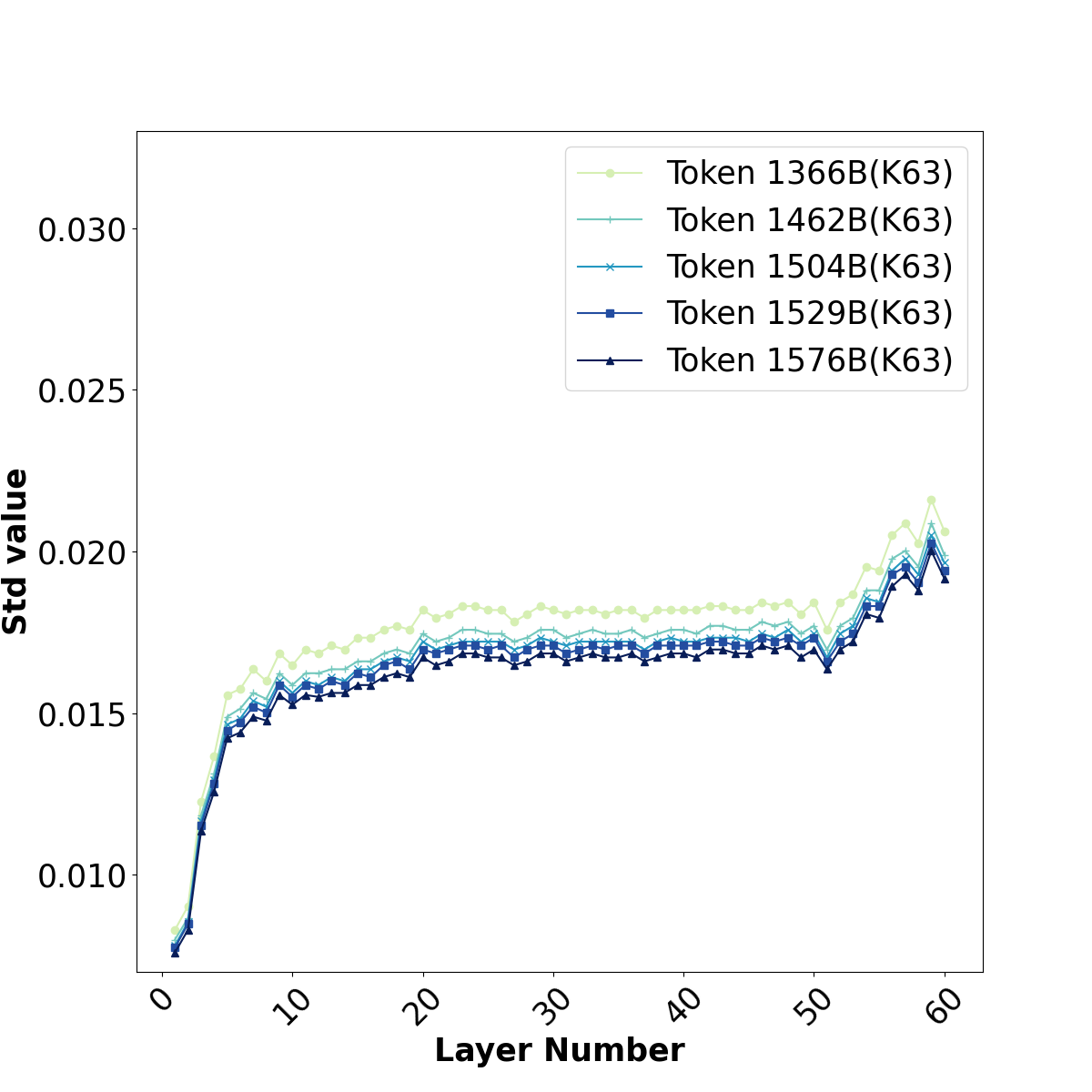}}
\end{subfigure}
\hfill
\begin{subfigure}[K64]{\includegraphics[width=0.48\linewidth]{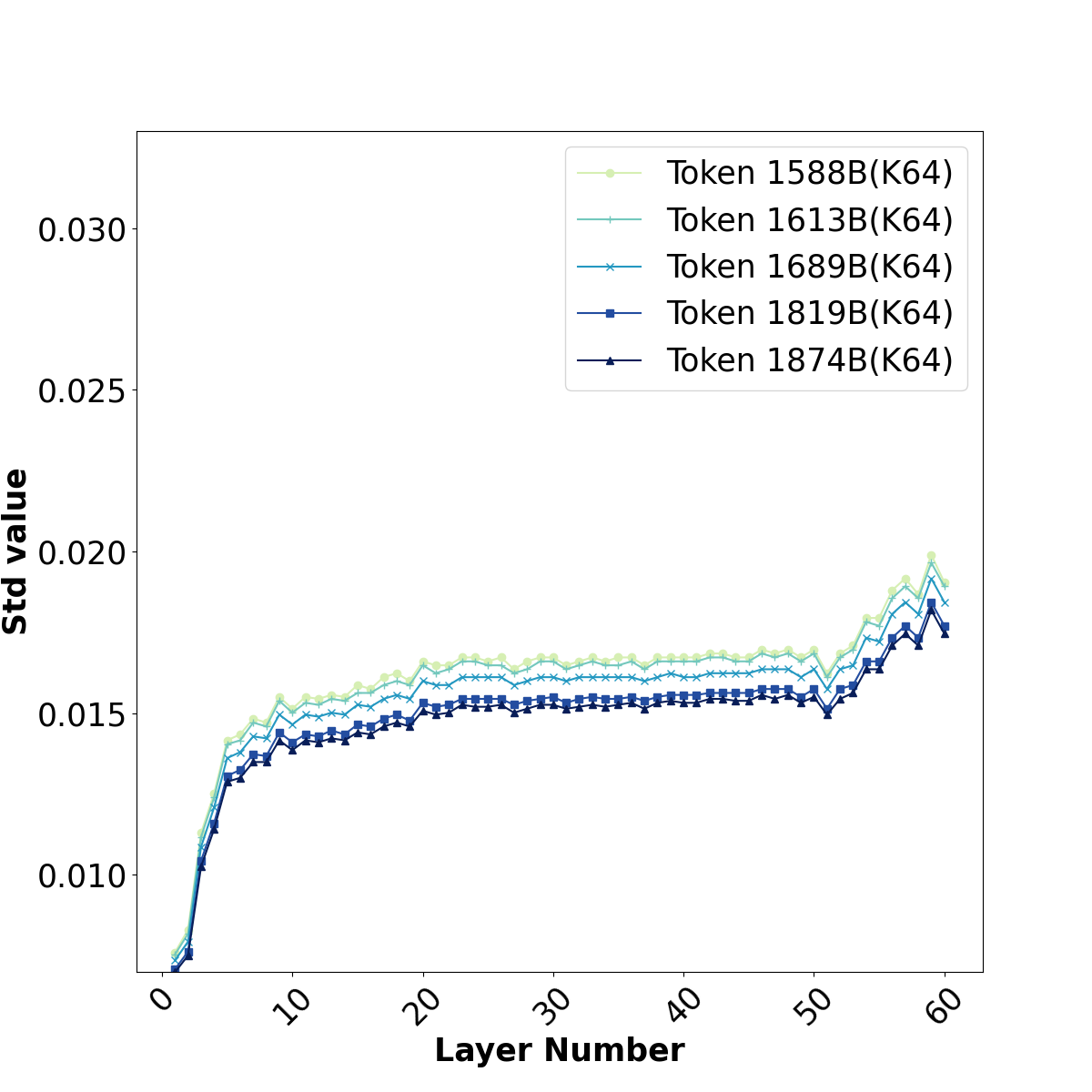}}
\end{subfigure}

\caption{Standard variance of the weight matrix $W_O$ across different layers in Aquila-34B. Different tokens in the graph represent the total amount of training data tokens used.}
\label{34B_o}
\end{figure*}

\begin{figure*}[hbpt]
\centering
\begin{subfigure}[K6]{\includegraphics[width=0.32\linewidth]{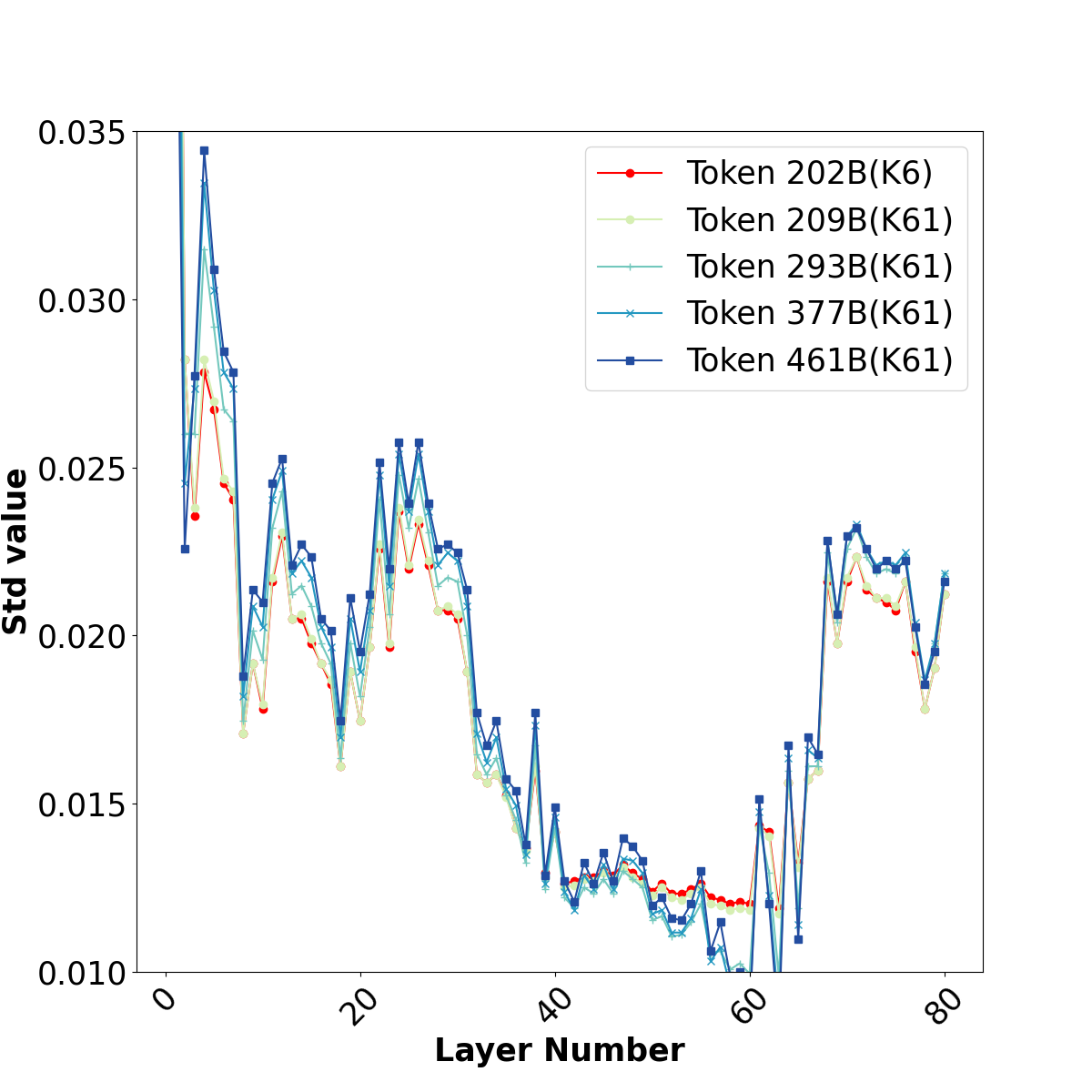}}
\end{subfigure}
\hfill
\begin{subfigure}[K63]{\includegraphics[width=0.32\linewidth]{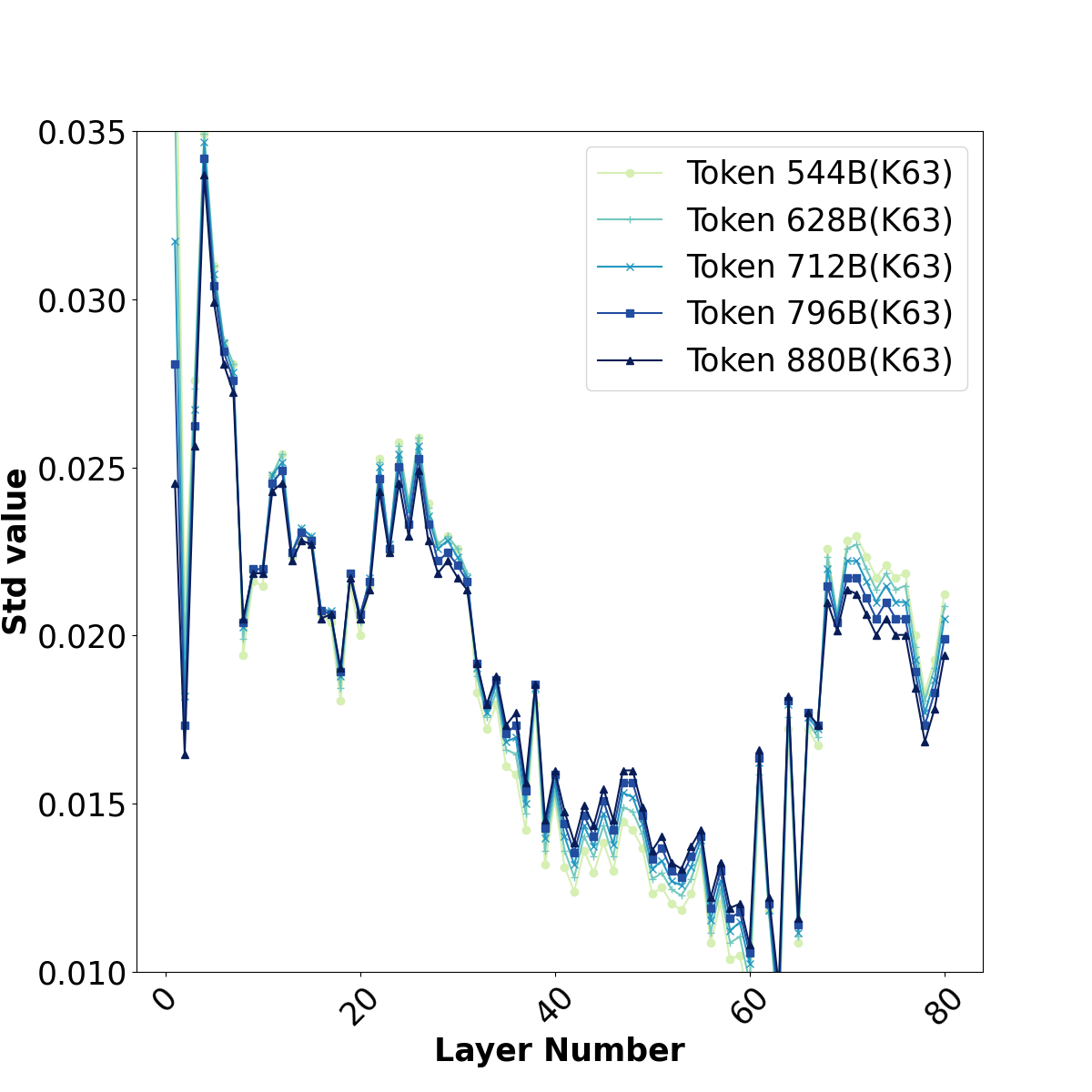}}
\end{subfigure}
\hfill
\begin{subfigure}[K65]{\includegraphics[width=0.32\linewidth]{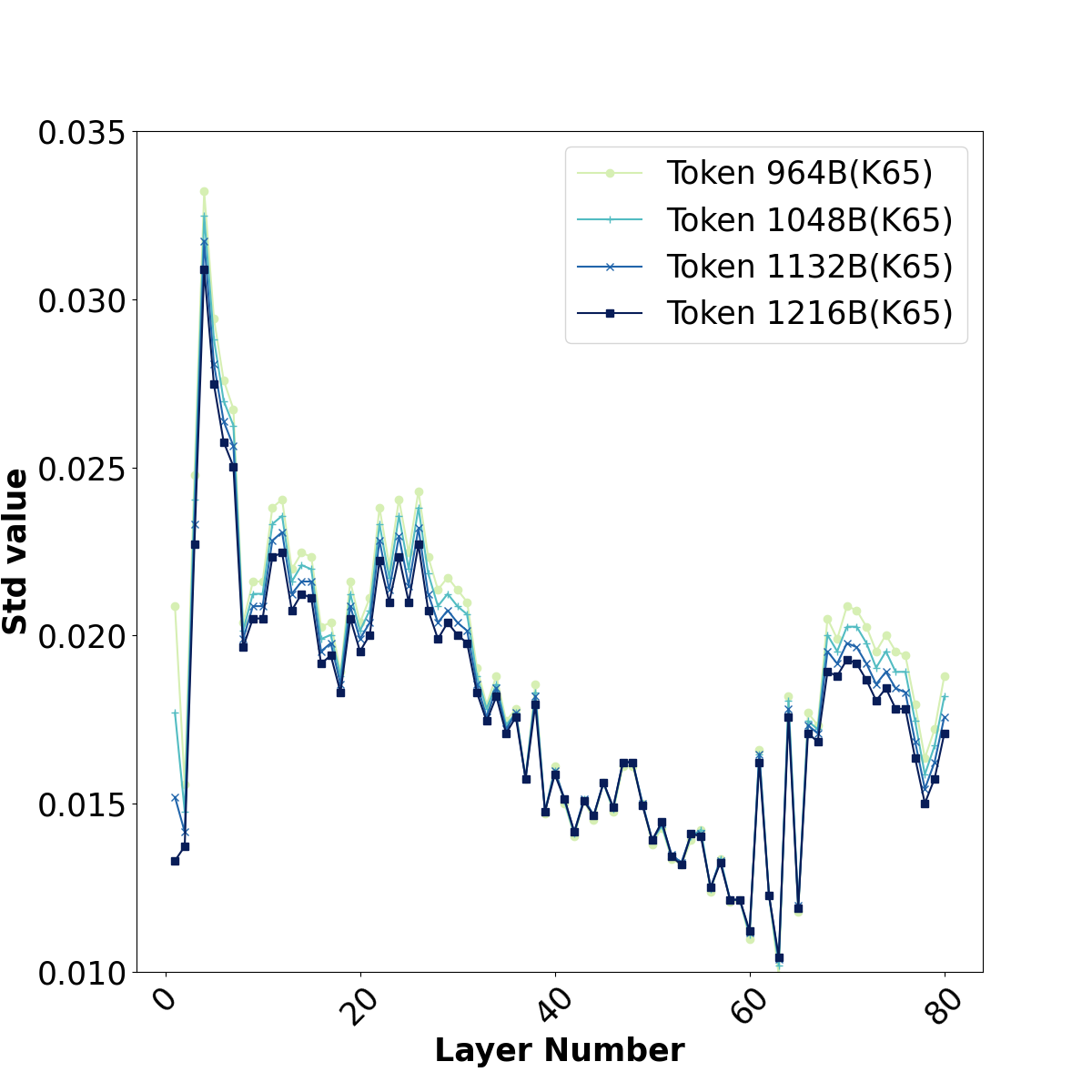}}
\end{subfigure}
\caption{Standard variance of the weight matrix $W_K$ across different layers in Aquila-70B. Different tokens in the graph represent the total amount of training data tokens used.}
\label{70B_k}
\end{figure*}

\begin{figure*}[hbpt]
\centering
\begin{subfigure}[K6]{\includegraphics[width=0.32\linewidth]{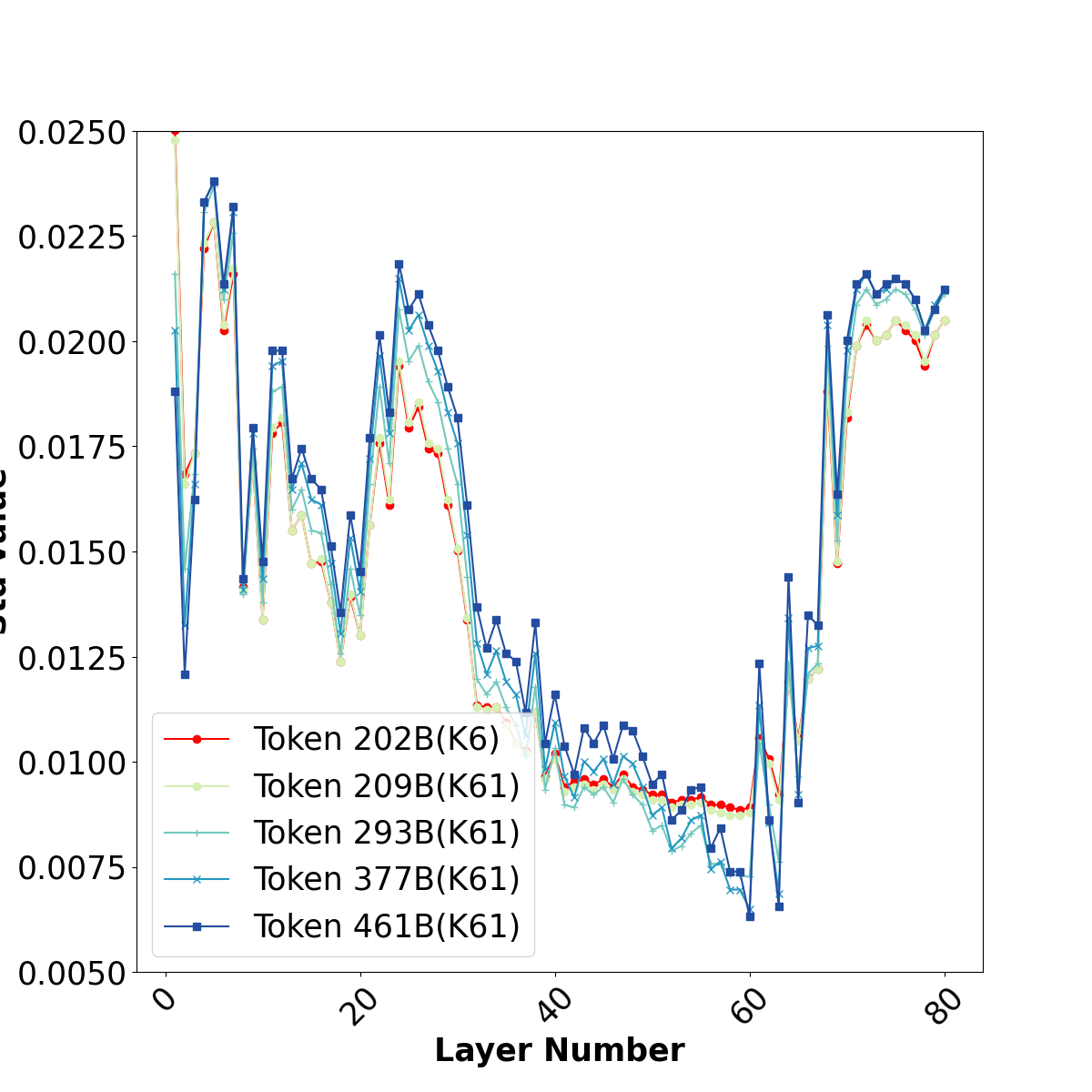}}
\end{subfigure}
\hfill
\begin{subfigure}[K63]{\includegraphics[width=0.32\linewidth]{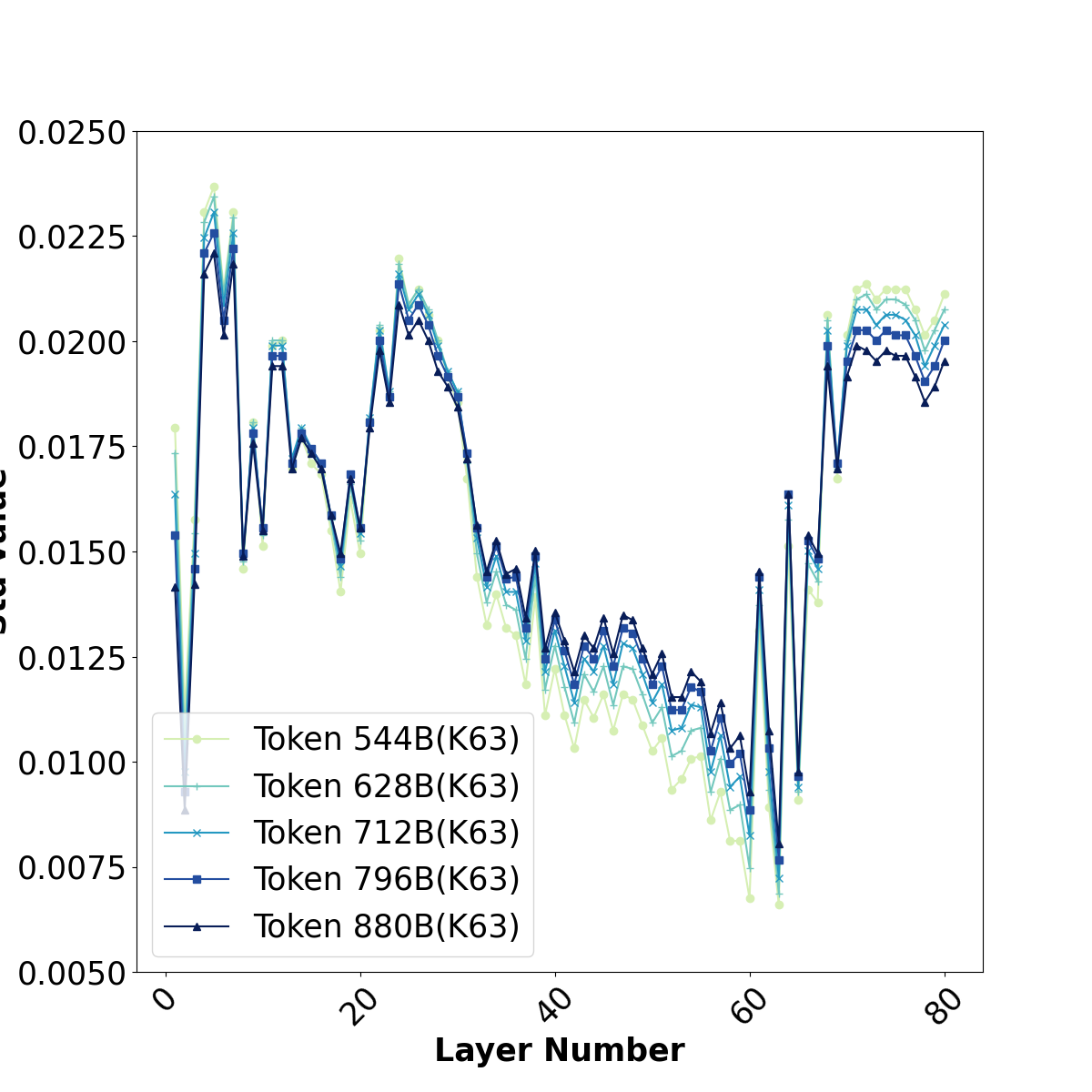}}
\end{subfigure}
\hfill
\begin{subfigure}[K65]{\includegraphics[width=0.32\linewidth]{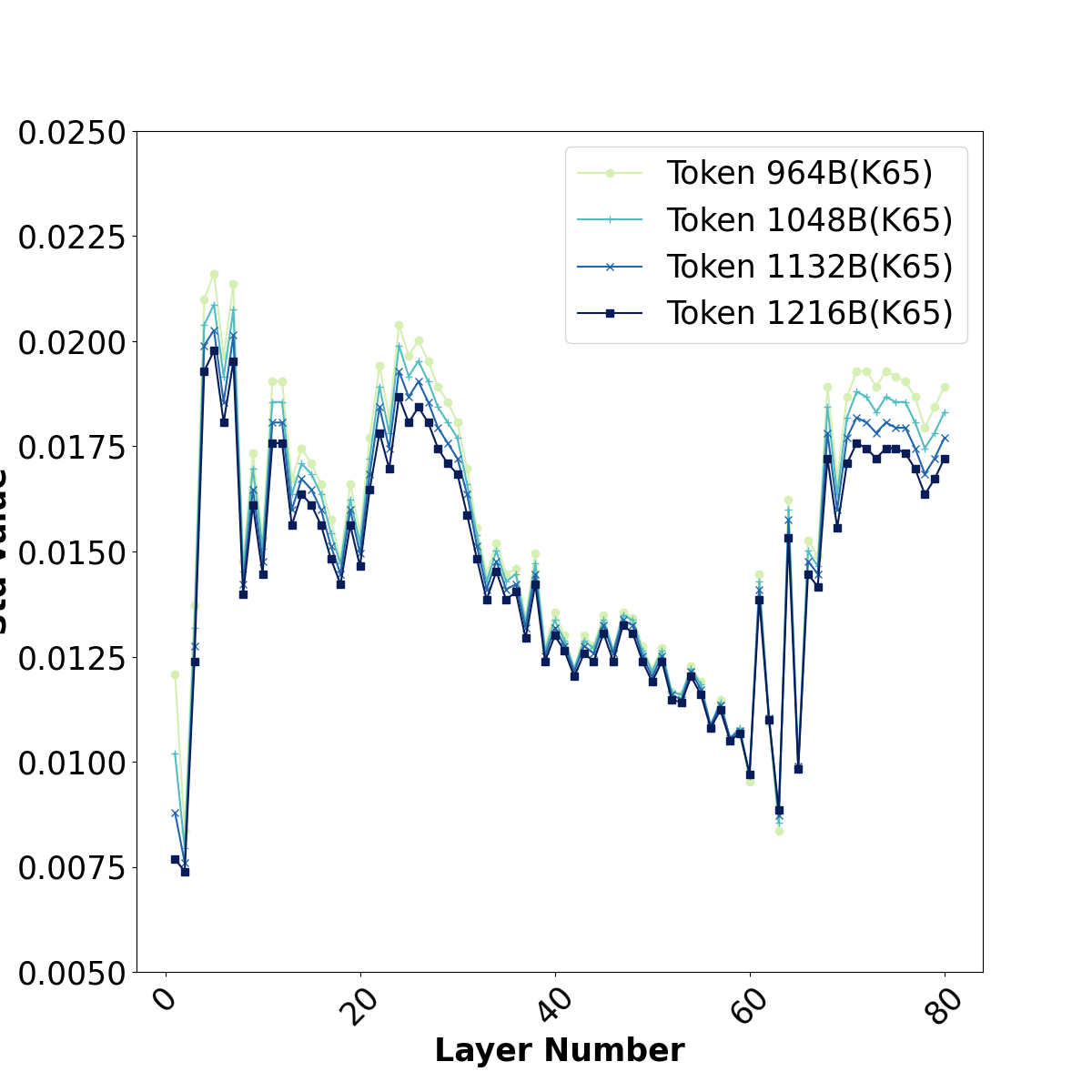}}
\end{subfigure}
\caption{Standard variance of the weight matrix $W_Q$ across different layers in Aquila-70B. Different tokens in the graph represent the total amount of training data tokens used.}
\label{70B_q}
\end{figure*}

\begin{figure*}[hbpt]
\centering
\begin{subfigure}[K6]{\includegraphics[width=0.32\linewidth]{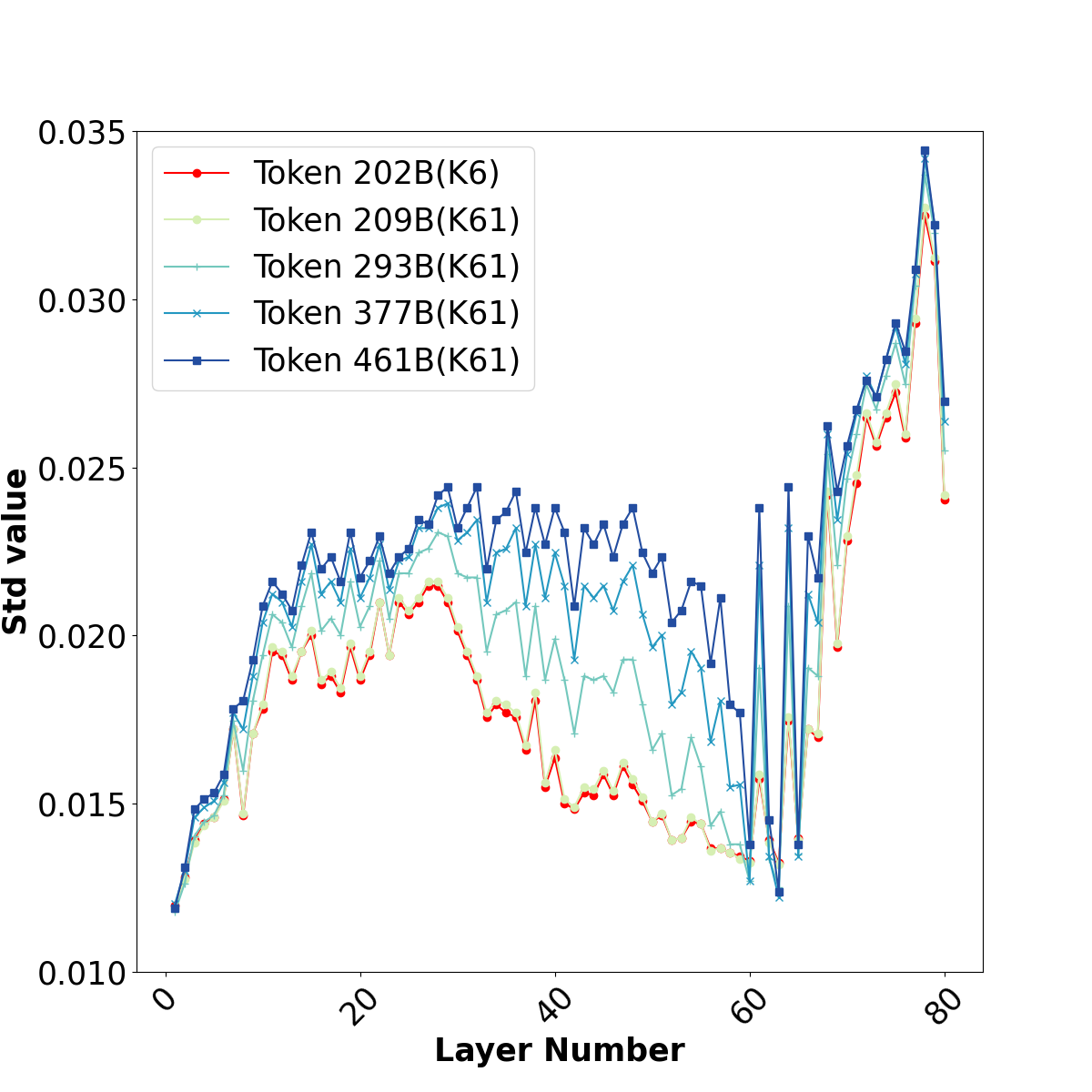}}
\end{subfigure}
\hfill
\begin{subfigure}[K63]{\includegraphics[width=0.32\linewidth]{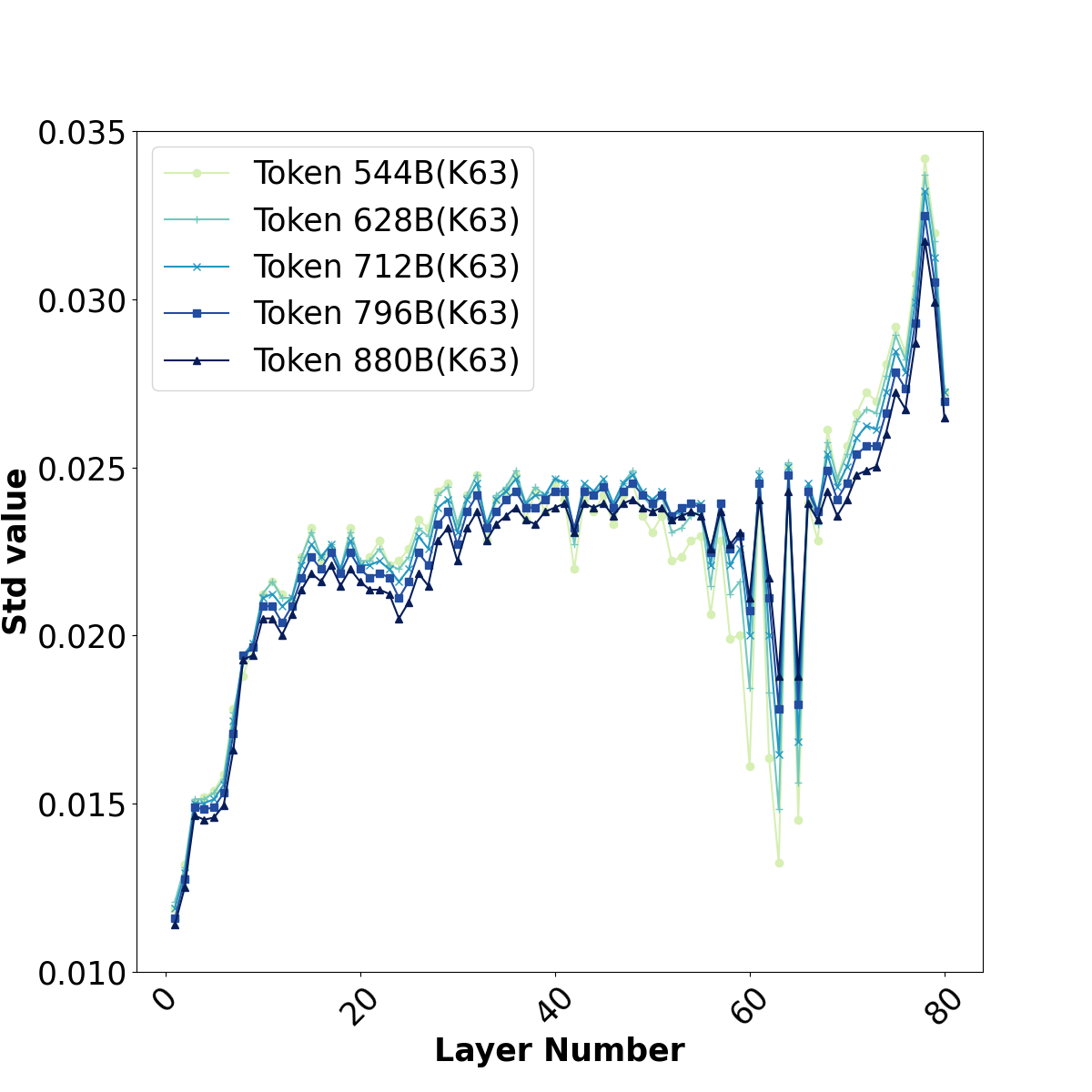}}
\end{subfigure}
\hfill
\begin{subfigure}[K65]{\includegraphics[width=0.32\linewidth]{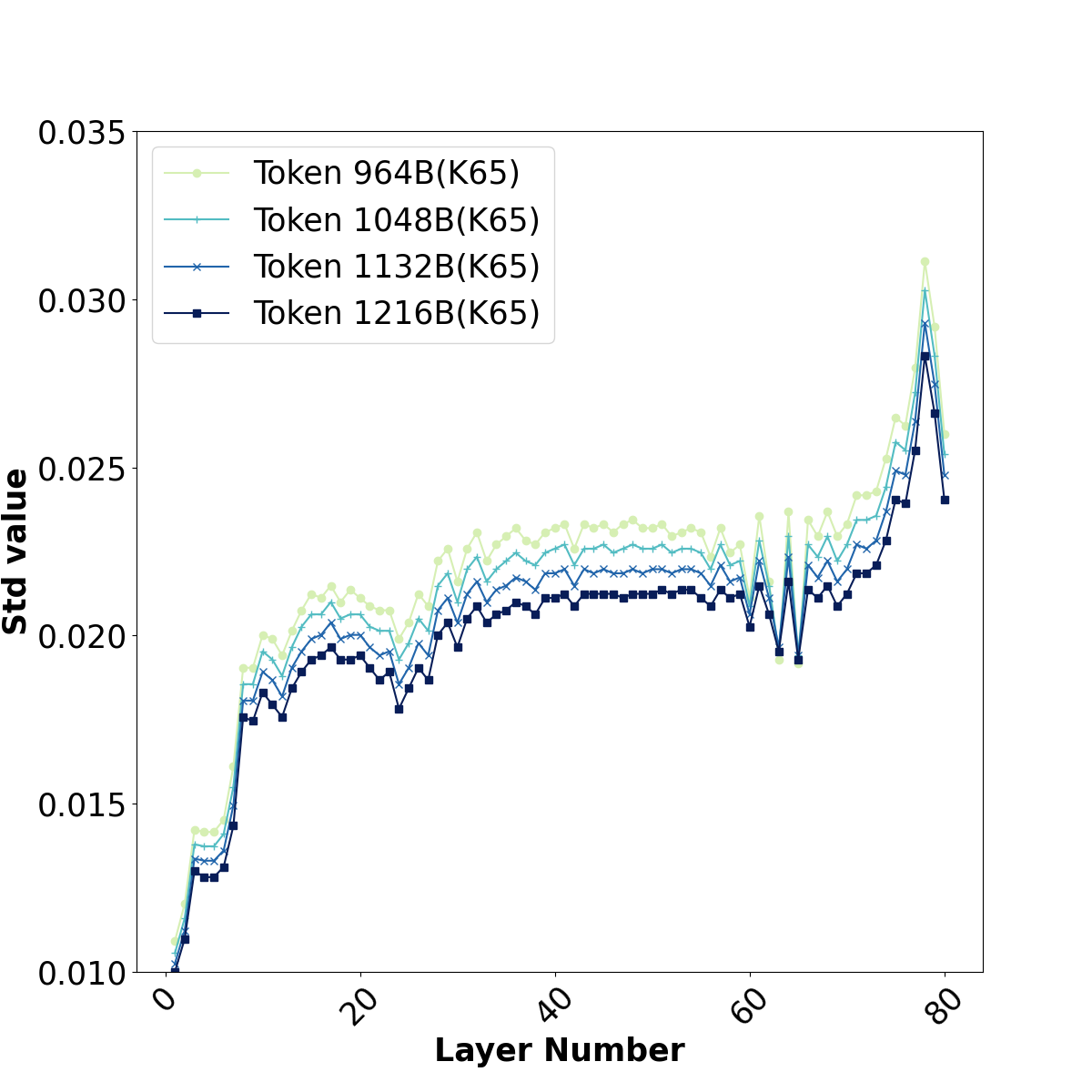}}
\end{subfigure}
\caption{Standard variance of the weight matrix $W_V$ across different layers in Aquila-70B. Different tokens in the graph represent the total amount of training data tokens used.}
\label{70B_v}
\end{figure*}

\begin{figure*}[hbpt]
\centering
\begin{subfigure}[K6]{\includegraphics[width=0.32\linewidth]{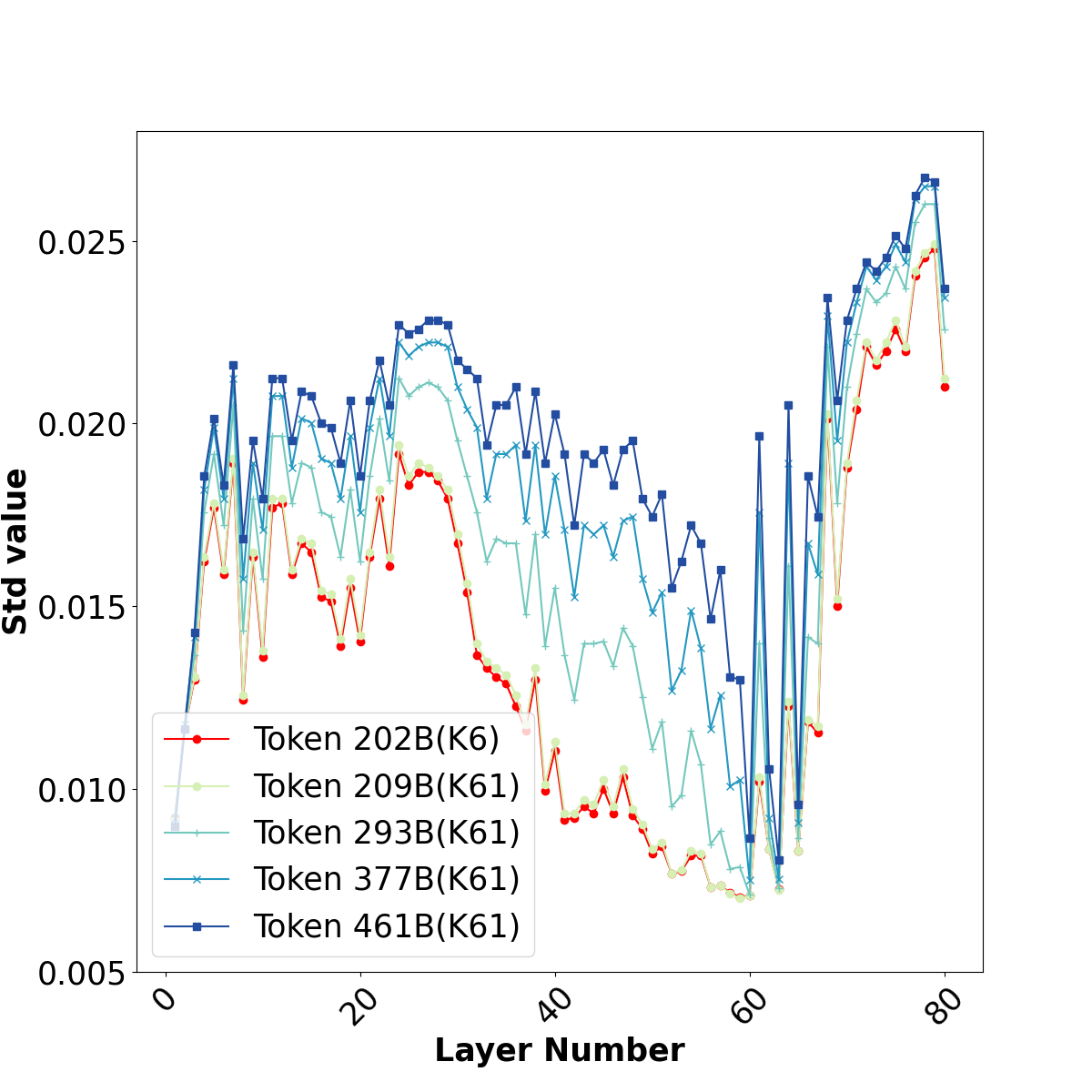}}
\end{subfigure}
\hfill
\begin{subfigure}[K63]{\includegraphics[width=0.32\linewidth]{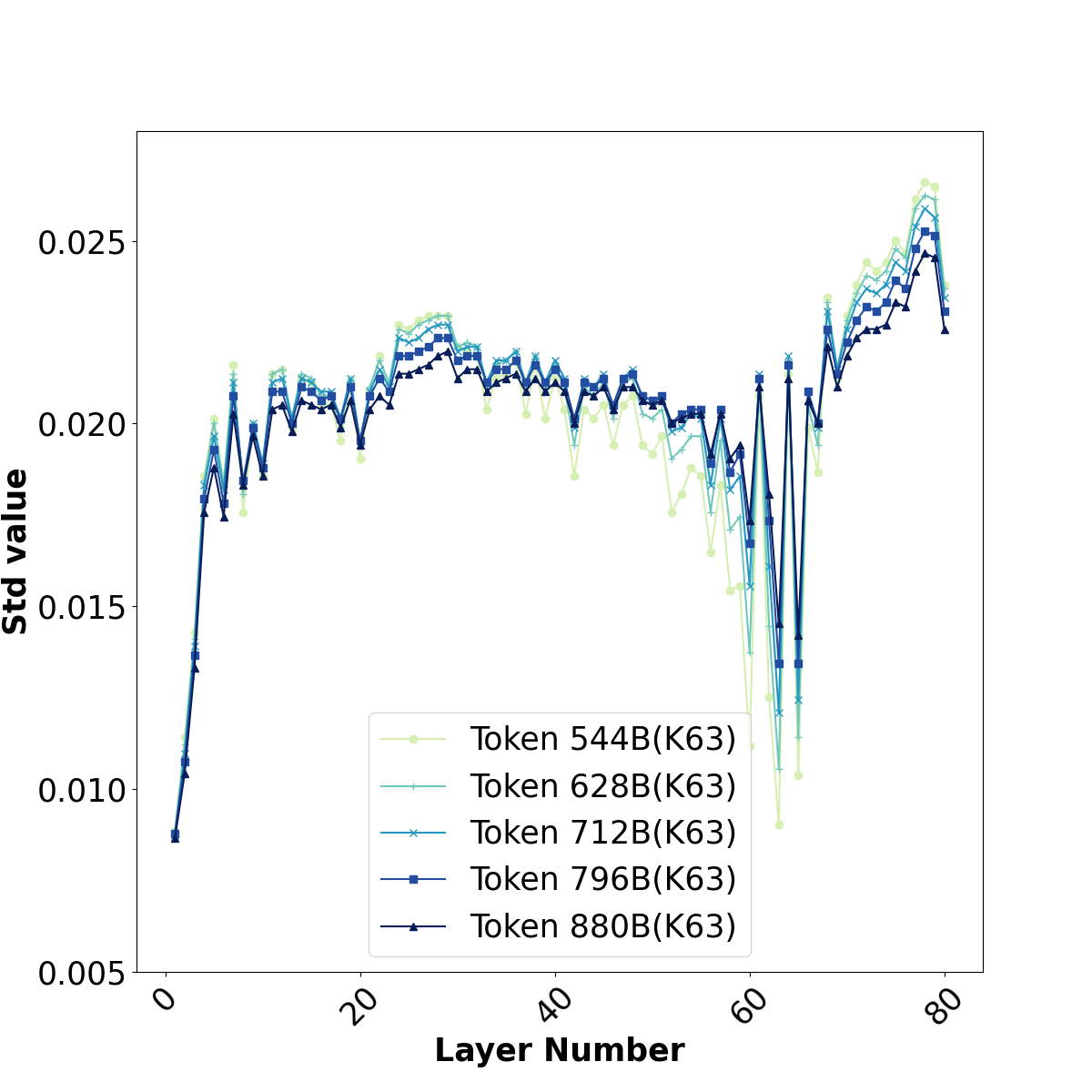}}
\end{subfigure}
\hfill
\begin{subfigure}[K65]{\includegraphics[width=0.32\linewidth]{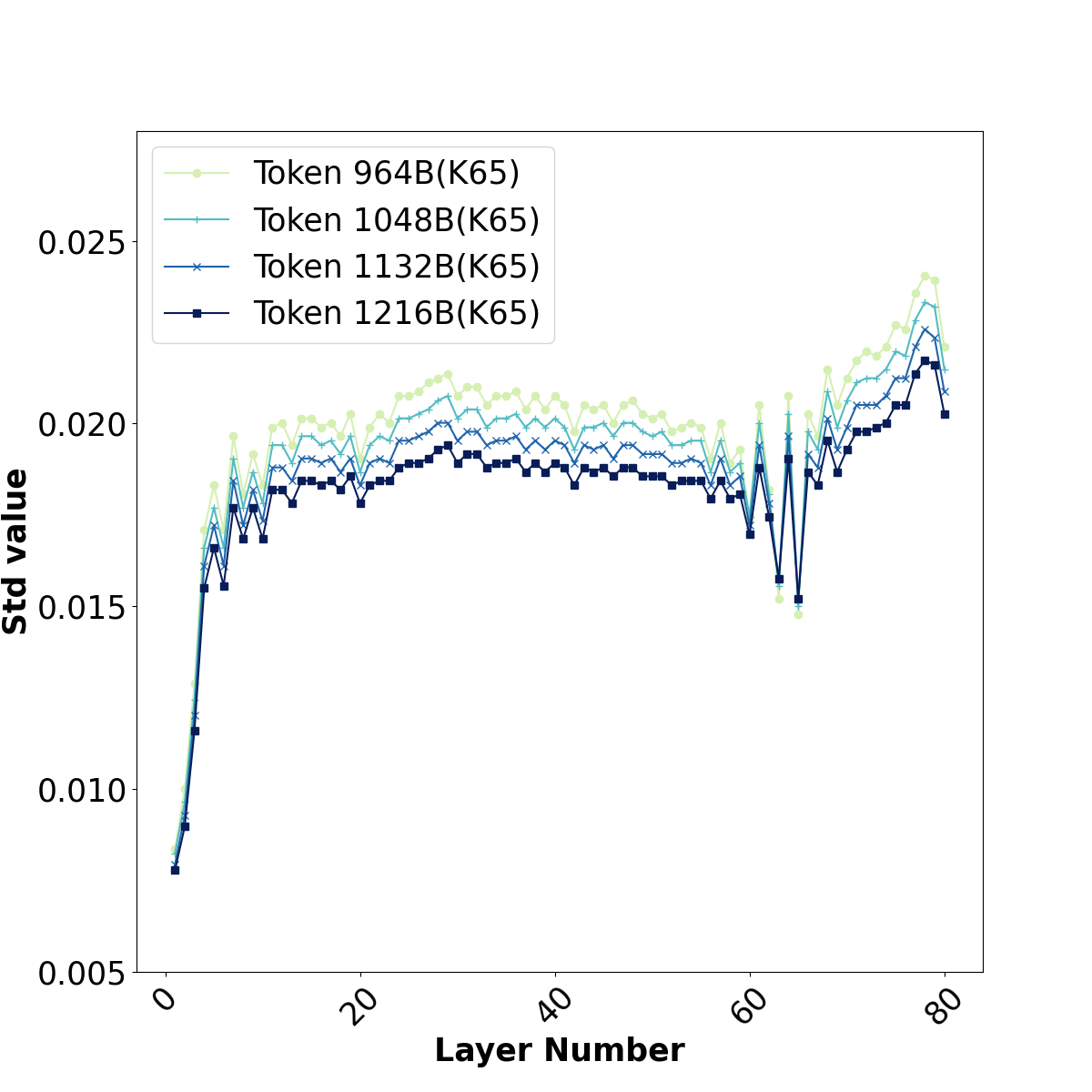}}
\end{subfigure}
\caption{Standard variance of the weight matrix $W_O$ across different layers in Aquila-70B. Different tokens in the graph represent the total amount of training data tokens used.}
\label{70B_o}
\end{figure*}

We plot the trajectories of the standard deviations of each parameter, namely self\_attn.k\_proj ($W_K$), self\_attn.q\_proj ($W_Q$), self\_attn.v\_proj ($W_V$), and self\_attn.o\_proj ($W_O$), for each layer, shown in Fig.~\ref{34B_k}, ~\ref{34B_q}, ~\ref{34B_v}, ~\ref{34B_o}, ~\ref{70B_k}, ~\ref{70B_q}, ~\ref{70B_v}, ~\ref{70B_o}. The y-axis represents the standard deviation, while the x-axis represents the index of each layer. Each trajectory serves as a representation of the corresponding parameter in a checkpoint. Each subplot represents a training stage, namely K6, K61\&K62, K63, K64 for Aquila2-34B, and K6, K61, K63, K65 for Aquila-70B.

\subsection{Detailed results of the Chat model.}
As illustrated in Fig.\ref{tab:detailed-chat}, AquilaChat2-34B-v1.2 and AquilaChat2-34B-v1.1 achieve the highest scores in most tasks. Moreover, AquilaChat2-34B-v1.2 achieves the highest average scores in both Chinese and English tasks.
\begin{table*}[ht]
\setlength{\tabcolsep}{2.5pt}
\centering
\begin{tabular}{ccccccccc}
Model           &\rotatebox{80}{AquilaChat2-34B-v1.2}    & \rotatebox{80}{AquilaChat2-34B-v1.1} &\rotatebox{80}{Baichuan2-13B-chat} & \rotatebox{80}{AquilaChat2-7B}  & \rotatebox{80}{Llama2-70B-chat}        & \rotatebox{80}{InternLM-7B-chat}     & \rotatebox{80}{Baichuan2-7B-chat}    & \rotatebox{80}{ChatGLM2-6B}     \\ \hline
Mean  &\textbf{76.34}   &74.49  &63.27  &63.21  &61.25   &60.84  &57.99  &34.13  \\
En Mean   &\textbf{69.92} &69.19  &59.09   &59.38   &65.13  &53.95  &55.54 &39.89  \\  
Zh Mean   &\textbf{79.19}  &77.43  &65.58  &65.04   &58.83  &66.48  &58.28 &28.75 \\   \hline
CLUEWSC(5-shot)    &\textbf{87.60} &87.32 &78.26 &56.86  &79.07 &75.27 &68.14 &50.70\\     
BUSTM(5-shot)      &82.50 &\textbf{84.44} &75.37 &72.80  &62.97 &76.13 &63.08 &-\\  
BoolQ (5-shot)      &86.80 &\textbf{88.00} &74.90 &78.10&88.57  &76.77 &68.47 &73.60 \\   
TruthfulQA(5-shot)    &64.80 &\textbf{70.49} &50.56 &66.21  &53.82 &33.28 &46.33 &27.52\\   
RAFT(5-shot)       &66.40 &\textbf{70.68} &74.24 &65.23  &70.91 &59.85 &66.82 &40.23 \\  
ChID(5-shot)        &\textbf{88.20} &87.78 &61.93 &72.80  &36.93 &68.43 &52.28 & 8.15 \\  
EPRSTMT(5-shot)    &90.00 &\textbf{91.64} &88.85 &90.98  &89.73 &89.95 &81.20 & 52.79\\   
TNEWS(5-shot)      &55.50 &\textbf{63.11} &47.68 &55.82  &45.04 &43.58 &39.48 & 4.37 \\    
OCNLI(5-shot)       &\textbf{83.60} &78.96 &50.48 &65.46  &50.62 &60.82 &51.15 &29.86\\   
SEM-Chinese(5-shot)   &\textbf{73.08} &55.74 &65.69 &44.17  &50.84 &59.29 &56.54 &35.86\\  
MMLU(5-shot)     &\textbf{61.70} &51.19 &52.96 &38.39  &55.14 &47.02 &48.31 & 18.20\\  
IRD    &58.30 &\textbf{65.60} &42.80 &49.00 & 57.20&52.80 &47.80 & - \\ 
CSL(5-shot)       &\textbf{76.40} &73.08 &48.76 &61.41  &55.42 &58.39 &54.36 & 48.27 \\  
CLCC-H-v2.0    &\textbf{75.80} & 74.84&73.24 &67.68  &- &50.16 &67.95 &54.17\\   \hline
\end{tabular}
\caption{Detailed results of the Chat model.}
\label{tab:detailed-chat}
\end{table*}

\subsection{Training parameters}
The parameters used for training Aquila2-34B and Aquila2-70B can be found in Tab.\ref{tab:aquila34b-config} and Tab.\ref{tab:aquila70b-config}, which include configuration information such as model structure, learning rate, optimizer, and so on. 
\label{app:training}
\subsection{Exploring Data Formulation Strategies in Specific Domains}\label{recipe}
\begin{itemize}
\item \textbf{v1 and v2 datasets} We refer to the sampling weights of various sources used in the GPT-3 work. A 2-3 epoch oversampling approach is employed for high-quality data like wiki and QA, while high-quality long textual data from textbooks and paper was oversampled for 1.5 epochs. For curated web data collections such as OpenWebText, the data were oversampled for 1.5 epochs, while other web data and code data were sampled for 1 epoch. Furthermore, in terms of language distribution, a balanced ratio of English to Chinese text to code was maintained at 60\%:30\%:10\%. The validation process was performed on the 7B model training. During the training process, challenges emerged due to significant variations in the data sources and disparate loss values during data fusion, leading to slow convergence of the model.

\item \textbf{v3 dataset} To address these issues, we considered the actual volume of data from each source. Web data was designated as the primary data source. Sampling weights for the data were determined based on the proportions of data obtained from various sources. This approach was inspired by the works of GPT-3~\cite{DBLP:conf/nips/BrownMRSKDNSSAA20} and Llama~\cite{DBLP:journals/corr/abs-2302-13971}. Specifically, the web data represented 82\%, with code, encyclopedia, and textbook data contributing 4.5\% each, literature 2.5\%, and questions and answers data 2\%. This refined methodology ensured a more effective and efficient training process, allowing improved convergence and overall model performance. 

\item \textbf{knowledge-oriented datasets (k1-k5)} Upon completion of training the 7B model on the v3 dataset (approximately 1000B tokens), we evaluated the model's performance with several validation datasets sampled from the same distribution to training datasets. Through the validation loss curve, we find that many knowledge-intensive datasets are not adequately learned. Therefore, we extracted pivotal subsets and collected richer, knowledge-enriched data to construct specialized knowledge-oriented datasets for continuous pretraining. Specifically, K1 dataset contains bilingual wiki datasets, bilingual paper datasets, English textbook datasets, and code-text pair datasets. In the k2 dataset, we introduce several new Chinese sources like question-and-answer forum data, technical blog data, newspaper data, and headline news data. The k3 dataset incorporates open-source literature data (such as redpajama~\cite{together2023redpajama}) and medical question-answering data. Additionally, the k4 dataset includes mathematical computation data and higher-quality open-source code data (from starcoder~\cite{li2023starcoder}). K5 further integrates a wider range of open-source high-quality Chinese data and data translated from sources (such as tigerbotdata~\url{https://huggingface.co/datasets/TigerResearch/pretrain_zh}, ccmatrix~\cite{schwenk2020ccmatrix}, etc.). During the training of the k1, k2, and k3 data sets, it was noted that the use of only knowledge-oriented data led to highly unstable model losses. To address this, in the k4 and k5 datasets, we supplemented knowledge-oriented data with the corresponding web data to ensure the stability of the model. Through experimentation with the k1 to k5 datasets, we validated the efficacy of high-quality data sources and determined the final configuration of the pretraining data set.
\end{itemize}

\end{document}